%% file: main.tex
\documentclass[twoside]{article}

\usepackage[preprint]{aistats2027}

\usepackage{mathrsfs}
\usepackage{hyperref}
\usepackage{url}
\usepackage{xurl}
\usepackage{enumitem}
\usepackage{graphicx}
\usepackage{amsmath}

\usepackage{bm}
\usepackage{amsfonts}
\usepackage{tikz}
\usepackage[dvipsnames,table]{xcolor}
\usepackage{hyperref}
\usepackage{cleveref}
\crefname{algorithm}{Algorithm}{Algorithms}
\Crefname{algorithm}{Algorithm}{Algorithms}
\crefname{assumption}{Assumption}{Assumptions}
\Crefname{assumption}{Assumption}{Assumptions}
\usepackage{nicefrac}       
\usepackage{microtype}      

\usepackage[most]{tcolorbox}
\usepackage{mleftright}
\usepackage[square,sort&compress]{natbib}

\hypersetup{
    colorlinks=true,       
    linkcolor=Red,         
    citecolor=blue,        
    filecolor=magenta,     
    urlcolor=Blue          
}

\usepackage{algorithm,caption}
\usepackage[noend]{algorithmic}

\usepackage{enumitem}

\usepackage{amsthm}

\newtheorem{definition}{Definition}[section]
\newtheorem{nad}{Notation and Definitions}[section]

\newtheorem{theorem}{Theorem}[section]
\newtheorem{lemma}[theorem]{Lemma}
\newtheorem{assumption}{Assumption}[section]

\usepackage{amssymb,booktabs,tabularx}

\usepackage[]{color-edits}
\addauthor[]{rd}{violet}
\addauthor[]{ug}{teal}
\addauthor[]{ks}{Green}
\addauthor[]{ab}{blue}

\input{notation}

\usepackage{tcolorbox}

\usepackage{url}

\begin{document}

\twocolumn[

\aistatstitle{Replicable Dueling Bandits}
\aistatsauthor{ Rohan Deb \And Aadirupa Saha \And Udaya Ghai }

\aistatsaddress{ University of Illinois, \\Urbana Champaign \\ \href{mailto:rd22@illinois.edu}{\texttt{rd22@illinois.edu}} \And
                 University of Illinois, Chicago \\ \href{mailto:aadirupa.saha@gmail.com}{\texttt{aadirupa.saha@gmail.com}} \And
                 Amazon \\ \href{mailto:ughai@amazon.com}{\texttt{ughai@amazon.com}} }

\aistatsauthor{ Karan Singh \And Arindam Banerjee }

\aistatsaddress{ Carnegie Melon University \\ \href{mailto:karansingh@cmu.edu}{\texttt{karansingh@cmu.edu}} \And
                 University of Illinois, Urbana Champaign \\ \href{mailto:arindamb@illinois.edu}{\texttt{arindamb@illinois.edu}} } ]

\begin{abstract}
  \input{section/abstract}

\end{abstract}

\section{Introduction}
\label{sec:intro}
\input{section/intro}

\section{Preliminaries}
\label{sec:preliminaries}
\input{section/preliminaries}

\section{Replicable Algorithms for Finite Dueling Bandits}
\label{sec:finite-dueling}
\input{section/repbtm}

\section{Replicable Linear Dueling Bandits}
\label{sec:repmaxinp}
\input{section/linear}

\section{Experiments}
\label{sec:exp}
\input{section/experiments}

\section{Conclusion}
\label{sec:conclusion}

\input{section/conclusion}

\subsection*{AI use statement}

In this work, we used generative AI tools to implement methods, support
qualitative and thematic data analysis, interpret results, generate synthetic
data sets, clean and reformat data sets, and propose or
refine hypotheses. We did not use generative AI tools to help develop theoretical models or conceptual frameworks, formulate
mathematical claims, provide critical ingredients for proving mathematical
claims, assist in the writing of proofs, design or provide feedback on research
methodology or experiments. Assistance with
translation is not applicable to this work. Additionally, we used generative AI
tools to create or modify scientific figures and to polish the writing of the
paper. We have reviewed all AI-assisted work: we checked LLM-generated research
ideas for potential plagiarism through a manual literature survey, and
LLM-generated code was verified and tested for correctness by one author. We
take responsibility for the final content of this work, including text, claims,
and artifacts produced with the aid of generative AI.


\bibliographystyle{plainnat}
\bibliography{references}

\clearpage
\appendix
\thispagestyle{empty}

\onecolumn
\aistatstitle{Replicable Dueling Bandits:
Supplementary Materials}

\section{Proofs for Finite Dueling Bandits}
\label{app:finite:proofs}

\input{appendix/repbtm}

\section{Complete Analysis of \textnormal{RepMaxInP}}
\label{app:repmaxinp}
\input{appendix/linear}

\section{Experimental Details}
\label{app:experiments}
\input{appendix/experimental_details}

\end{document}

%% file: notation.tex
\newcommand{\beq}{\begin{equation}}
\newcommand{\eeq}{\end{equation}}

\theoremstyle{definition} 

\newtheorem{remark}{Remark}[section]

\newcommand {\commentout}[1] {}

\def\ints{{{\rm Z} \kern -.35em {\rm Z} }}  
\def\smallints{{{\rm Z} \kern -.3em {\rm Z} }}  
\def\pints{{{\rm I} \kern -.15em {\rm N} }}      
\newcommand{\reals}{\mathbb R}

\def\cplx{{{\rm I} \kern -.45em {\rm C} }}       
\def\l2{\rm {\mathcal L}^{2}(\reals)}            

\newcommand{\be}{\begin{eqnarray}}
\newcommand{\ee}{\end{eqnarray}}
\newcommand{\bea}{\begin{eqnarray}}
\newcommand{\eea}{\end{eqnarray}}
\newcommand{\beaa}{\begin{eqnarray*}}
\newcommand{\eeaa}{\end{eqnarray*}}
\newcommand{\bnad}{\begin{nad}}
\newcommand{\enad}{\end{nad}}

\newcommand{\IGNORE}[1]{}

﻿

{\begin{array}{l@{\hspace{8mm}}l@{\hspace{8mm}}l}}%
{\end{array}}
\newlength{\lplb}
\DeclareMathSymbol{\mhyphen}{\mathord}{AMSa}{"39}

\newcommand{\RepBTM}{\color{Green}\ensuremath{\mathtt{RepBTM}}}

\newcommand{\RepBTMSST}{\color{Green}\ensuremath{\mathtt{RepBTM}\mhyphen \mathtt{SST}}}

\newcommand{\RepMaxInP}{\color{Green}\ensuremath{\mathtt{RepMaxInP}}}


%% file: section/abstract.tex
Dueling bandits choose each comparison based on earlier noisy outcomes, so two runs on the same preferences can compare different pairs. We study replicable dueling bandits, where two executions with shared internal randomness and independent comparison outcomes must select the same ordered pair in every round with probability at least \(1-\rho\). Building on Beat-the-Mean, whose scores change as arms are eliminated and need not rank the best arm highest, we design the first such algorithms with sublinear regret. Under relaxed stochastic transitivity with parameter \(\gamma\ge1\) and the stochastic triangle inequality, we design an algorithm that achieves expected regret \(\widetilde{\mathcal O}(\gamma^8K^4H_{\mathrm p}/\rho^2)\), where \(H_{\mathrm p}=\sum_{i=2}^{K}(P_{1i}-1/2)^{-1}\). Under strong stochastic transitivity, we design an algorithm based on simultaneous elimination that improves $K$-dependence and achieves \(\widetilde{\mathcal O}(K^2H_{\mathrm p}/\rho^2)\) regret. For linear dueling bandits, we design an algorithm that achieves regret \(\widetilde{\mathcal O}(d^3\sqrt{T}/(c_\mu\rho))\). Experiments on synthetic and language-model preferences show replicability, with sublinear regret.
\vspace{-1ex}

%% file: section/intro.tex
Learning from relative feedback \citep{BradleyTerry,furnkranz2010preference,wirth2017survey,bengs2021preference} has become a central paradigm in machine
learning.
In many applications, absolute ratings are noisy and inconsistent across
annotators, whereas pairwise judgments are cheaper to elicit and more reliable.
Preference-based learning has accordingly been applied to the online evaluation
of retrieval systems \citep{joachims2002optimizing,radlinski2008how,yue2009interactively,chapelle2012large,hofmann2016online},
recommendation \citep{jones2011comparisons,kalloori2016pairwise},
human-in-the-loop optimization \citep{brochu2007active,gonzalez2017preferential,tucker2020preference},
and reinforcement learning from human feedback \citep{christiano2017deep}, which
is now standard for fine-tuning large language models
\citep{ziegler2019fine,stiennon2020learning,ouyang2022training,bai2022training,rafailov2023direct}.
The dueling bandit framework \citep{joachims2007db,yue2012karmed} formalizes the
online version of this setting. Given $K$ arms, at round $t$ the learner
selects $(a_t, b_t)$ and observes a binary outcome
$o_t \sim \mathrm{Bernoulli}(P(a_t \succ b_t))$, with the goal of minimizing
regret with respect to the best arm.

A dueling bandit algorithm chooses each pair based on the outcomes of earlier
duels. This adaptivity makes it sample efficient, but it also makes the sequence
of comparisons random: two runs on the same preference instance can disagree on
a few early outcomes, favor different arms, and spend the rest of their budget
comparing different pairs. In the applications that motivate dueling bandits,
these pairs determine what is measured. In online evaluation of retrieval
systems, they determine which rankers are interleaved for users and therefore
which ranker is declared better \citep{yue2009interactively}. In active
preference collection for language models, they determine which responses
annotators compare and therefore the dataset on which the reward model is
trained \citep{das2025active}. If rerunning the algorithm on a fresh sample of
(i.i.d.) users yields different comparisons, its conclusions cannot be reproduced or
audited. Preference feedback makes this divergence likely. Each duel reveals a single
noisy bit, and human judgments are inconsistent: labelers in
\citet{ouyang2022training} agreed with each other on about 73\% of
comparisons. Moreover, under the BTL model \citep{BradleyTerry,luce2005individual}, a learner with low regret must
spend most rounds comparing near-optimal arms, whose preference probabilities
are close to $1/2$, so the outcomes that drive its choices are close to
coin flips. This raises the question of whether a dueling bandit algorithm can
make the same comparisons across runs while still achieving low regret.

\citet{impagliazzo2022reproducibility} formalized this requirement as
algorithmic \emph{replicability}. A randomized algorithm is $\rho$-replicable if two
executions that share their internal randomness, but receive independent
samples from the same distribution, return the same output with probability at
least $1-\rho$. Replicability has since been studied for PAC learning,
clustering, and statistical estimation
\citep{bun2023stability,kalavasis2023statistical,chase2023replicability,esfandiari2023clustering}.
In sequential decision making replicable algorithms have been developed for
stochastic multi-armed and linear bandits
\citep{esfandiari2023replicable,komiyama2024replicability,chen2025regret,deb2026replicable},
constrained bandits \citep{bollini2026replicable}, reinforcement learning
\citep{karbasi2023replicability,eaton2023replicable}, and online learning
\citep{ahmadi2025replicable}. \citet{zhang2025replicable} study the
replicability of statistical analyses of digital health trials that assign treatments with bandit algorithms. In all of these settings, the learner observes a numerical reward or loss for the action it takes. Replicability under preference feedback, where the learner observes only the outcome of a comparison between two arms, has not been studied. We study this requirement for dueling bandits. A learner is \(\rho\)-replicable if two executions on the same preference
instance, using a shared internal seed and independent sources of comparison randomness, select the same ordered pair at every round with probability at least \(1-\rho\). Our objective is to achieve this agreement along with low cumulative regret.

\begin{table*}[t]
\centering
\small
\setlength{\tabcolsep}{4pt}
\renewcommand{\arraystretch}{1.16}
\caption{Representative regret guarantees for \(\rho\)-replicable
stochastic bandits.
For reward bandits, \(H_{\mathrm{r}}=
\sum_{i:\mu_i<\mu_\star}(\mu_\star-\mu_i)^{-1}\);
for finite dueling bandits, \(H_{\mathrm{p}}=
\sum_{i=2}^{K}(P_{1i}-1/2)^{-1}\).
Finite dueling rows assume a strict preference order and the stochastic
triangle inequality.
Bounds are in expectation; the linear dueling row uses utility regret.
\(\widetilde{\mathcal O}\) suppresses polylogarithmic factors, and
\(c_\mu\) is a positive lower bound on the link derivative.}
\label{tab:related}
\begin{tabularx}{\textwidth}{@{}
  >{\raggedright\arraybackslash}p{0.21\textwidth}
  >{\raggedright\arraybackslash}p{0.25\textwidth}
  >{\raggedright\arraybackslash}X
  l@{}}
\toprule
Work & Setting & Approach & Regret \\
\midrule
\citet{esfandiari2023replicable}
& \(K\)-armed rewards
& Batched elimination
& \(\displaystyle
  \mathcal O\!\left(
  \frac{K^2H_{\mathrm{r}}\log T}{\rho^2}\right)\) \\
\citet{komiyama2024replicability}
& \(K\)-armed rewards
& Batched elimination
& \(\displaystyle
  \widetilde{\mathcal O}\!\left(
  H_{\mathrm{r}}\Big(
  \log T+\frac{K^2}{\rho^2}\Big)\right)\) \\
\citet{deb2026replicable}
& \(K\)-armed rewards
& Batched optimism
& \(\displaystyle
  \widetilde{\mathcal O}\!\left(
  \frac{K^2H_{\mathrm{r}}\log T}{\rho^2}\right)\) \\
\midrule
\rowcolor{blue!8}
\textbf{Ours}
& \mbox{\(K\)-armed dueling, \(\gamma\geq1\)}
& Single-arm elimination
& \(\displaystyle
  \widetilde{\mathcal O}\!\left(
  \frac{\gamma^8K^4H_{\mathrm{p}}}{\rho^2}\right)\) \\
\rowcolor{blue!8}
\textbf{Ours}
& \mbox{\(K\)-armed dueling, \(\gamma=1\)}
& Simultaneous elimination
& \(\displaystyle
  \widetilde{\mathcal O}\!\left(
  \frac{K^2H_{\mathrm{p}}}{\rho^2}\right)\) \\
\midrule
\citet{deb2026replicable}
& Linear rewards
& Batched optimism
& \(\displaystyle
  \widetilde{\mathcal O}\!\left(
  \frac{d^3}{\rho}\sqrt{T}\right)\) \\
\rowcolor{blue!8}
\textbf{Ours}
& Linear dueling
& Batched maximum-informative-pair
& \(\displaystyle
  \widetilde{\mathcal O}\!\left(
  \frac{d^3}{c_\mu\rho}\sqrt{T}\right)\) \\
\bottomrule
\end{tabularx}
\end{table*}

\textbf{Contributions.} The primary contribution of this work is a theoretical treatment of trajectory replicability in stochastic dueling bandits.
Our main results concern two nested classes of preferences over \(K\) arms.
The first permits relaxed stochastic transitivity with parameter
\(\gamma\geq1\); the second imposes strong stochastic transitivity,
corresponding to \(\gamma=1\).
Both also assume the stochastic triangle inequality. See Section~\ref{sec:preliminaries} for details.
We obtain gap-dependent regret guarantees in both regimes and improve the
polynomial dependence on \(K\) by a factor of \(K^2\) under the stronger
assumption. We also give a replicable algorithm for linear dueling bandits with sublinear regret. 
Table~\ref{tab:related} summarizes these results alongside representative reward-bandit guarantees and our main contributions are as follows.
\vspace{-2ex}
\begin{enumerate}
\item \textbf{Replicable dueling bandits under relaxed transitivity.} We give a \(\rho\)-replicable algorithm  for \(K\)-armed dueling bandits \(\RepBTM\), under relaxed stochastic transitivity with parameter \(\gamma\ge1\) and stochastic triangle inequality. Its regret is \(\widetilde{\mathcal O}(\gamma^8K^4H_{\mathrm p}/\rho^2)\), where \(H_{\mathrm p}\!=\!\sum_{i=2}^K\epsilon_i^{-1}\) and \(\epsilon_i=P_{1,i}-\frac{1}{2}\) is the gap of arm \(i\) to the best arm.
\item \textbf{Closing the gap under strong transitivity.} Under strong stochastic transitivity (\(\gamma=1\)), we give \(\RepBTMSST\), whose expected regret is \(\widetilde{\mathcal O}(K^2H_{\mathrm p}/\rho^2)\). This matches the \(K^2/\rho^2\) replicability overhead of elimination algorithms for reward bandits \citep{esfandiari2023replicable}.
\item \textbf{Linear preferences.} For preferences generated by \(d\)-dimensional linear utilities through a known increasing link, we give a \(\rho\)-replicable algorithm $\RepMaxInP$. With high probability, its regret is \(\widetilde{\mathcal O}(d^3\sqrt T/(c_\mu\rho))\), where \(c_\mu\) lower-bounds the derivative of the link.
\item \textbf{Experiments.} 
We supplement these theoretical results with numerical illustrations on synthetic and language-model preferences. These illustrations show that non-replicable baselines diverge early, whereas our algorithms keep the fraction of identical trajectories near the target \(1-\rho\) and still commit to the best arm.
\end{enumerate}

\textbf{Challenges and techniques.} With preference feedback, an arm's quality is defined only relative to its opponents, so BTM-style algorithms estimate the score \(p_i(W)=|W|^{-1}\sum_{j\in W}P_{i,j}\) of each arm against the active set \(W\). Unlike mean rewards, these targets change whenever an arm is eliminated. BTM also reuses comparisons across eliminations, so the active set depends on the very samples used to estimate its scores. We show that each retained record list is a prefix of a fixed sampling sequence filtered to the active set, which gives concentration uniformly over all candidate sets. For replicability, we round estimates on a shared random grid and couple both executions with a reference execution that uses rounded population scores, whose active sets depend only on the rounding offsets.

Relaxed transitivity adds a second difficulty: the best arm need not maximize \(p_i(W)\), so \(\RepBTM\) removes only one minimum-score arm at a time. With reused records, this requires a separate rounded estimate for every arm, stage, and sample level, \(\mathcal O(K^2\log T)\) in total, which leads to a \(K^4/\rho^2\) overhead. Replicable reward-bandit algorithms pay only \(K^2/\rho^2\), and without replicability BTM has the same dependence on \(K\) as reward-bandit algorithms \citep{yuejoachims2011}, so the gap is due to replicability.

Under strong transitivity, best arm maximizes \(p_i(W)\) in every active set containing it, so \(\RepBTMSST\) can eliminate several arms at once. We use phases with doubling budgets and fresh samples, which reduces the number of rounded estimates to \(\mathcal O(K\log T)\). The key step is bounding the exploration cost: an arm's score gap can be as small as \(\epsilon_i/|W|\), and a per-arm analysis would lose another factor of \(K^2\). Our rank-sensitive survivor bound avoids this loss, since an arm with a large gap can survive a phase only if the active set shrinks proportionally.

\textbf{Open Problem:} Whether the \(K^2/\rho^2\) overhead is achievable for \(\gamma>1\) and what the optimal dependence on \(K\) is, remain open (Section~\ref{sec:conclusion}).

%% file: section/preliminaries.tex
We consider stochastic dueling bandits with a known horizon \(T\).
Our main setting consists of a fixed set of \(K\) arms,
\(\mathcal A=\{b_1,\dots,b_K\}\).
At each round \(t\in[T]\), the learner selects an ordered pair
\((i_t,j_t)\in[K]\times[K]\) and observes a binary outcome
\(O_t\), where \(O_t=1\) indicates that \(b_{i_t}\) wins.
Self-comparisons are allowed.

An instance is specified by an unknown preference matrix
\(P\in[0,1]^{K\times K}\), where
\(P_{i,j}+P_{j,i}=1\) and \(P_{i,i}=\tfrac12\).
Let \(\mathcal F_t^{-}\) contain the interaction history,
the algorithm's randomness revealed so far, and the selected pair,
immediately before observing \(O_t\).
The feedback satisfies
\[
  \Pr(O_t=1\mid\mathcal F_t^{-})=P_{i_t,j_t},
\]

with fresh comparison randomness at each round.
Write \(\epsilon_{i,j}:=P_{i,j}-\tfrac12\), so that \(\epsilon_{i,j}=-\epsilon_{j,i}\).

\textbf{Best arm and regret.}
The preference assumptions introduced in section~\ref{sec:finite-dueling} ensure a unique best arm:
an arm that beats every other arm with probability strictly greater
than \(\tfrac12\).
For analysis, we label this arm \(b_1\); its identity is unknown
to the learner.
Define its preference gaps by
\(\epsilon_i:=\epsilon_{1,i}\), so that
\(\epsilon_1=0\) and \(\epsilon_i>0\) for \(i\geq2\).
Following \citet{yuejoachims2011}, we measure performance by
the cumulative probability regret $R_T$:
\begin{align*}
  R_T:=\sum_{t=1}^{T}r_t,
  \qquad r_t:=\frac12\bigl(\epsilon_{i_t}+\epsilon_{j_t}\bigr).
\end{align*}
Note that \(0\leq r_t\leq\tfrac12\), and comparing the best arm
with itself incurs zero regret.
We write \(H_{\mathrm p}:=\sum_{i=2}^{K}\epsilon_i^{-1}\)
for the gap-dependent complexity.

\textbf{Replicability.}
We compare two executions on the same instance.
They share the algorithm's internal random seed but use independent
sources of comparison randomness, independent also of the shared seed.

\begin{definition}[Trajectory replicability]
\label{def:replicability}
Fix \(\rho\in(0,1)\).
An algorithm is \(\rho\)-replicable if, for every admissible instance,
its two executions satisfy
\begin{align*}
  \Pr\!\big(
    (i_t^{(1)},j_t^{(1)})=(i_t^{(2)},j_t^{(2)})
    \text{ for every }t\in[T]
  \big)
  \geq1-\rho.
\end{align*}
The probability is over the shared internal seed and the independent comparison randomness.
\end{definition}

This condition concerns the entire sequence of ordered comparisons.
We use \(\delta\) for statistical failure probability and
\(\rho\) for the replicability failure probability.

\textbf{Linear Setting.}
For the linear setting, we allow a varying finite indexed action set
\(\mathcal A_t=\{a_{t,1},\dots,a_{t,K_t}\}\), where \(K_t\geq1\).
Each action has an observed feature vector
\(x_{t,i}\in\mathbb R^d\).
The linear preference model and the regret are specified
in Section~\ref{sec:repmaxinp}.


%% file: section/repbtm.tex
Our finite-arm results rely on the following assumptions on the preference
matrix, which follow \citet{yue2012karmed} and \citet{yuejoachims2011}.

\begin{assumption}[Strict preference order]
\label{asmp:strict-order}
There is an ordering \(b_1\succ b_2\succ\cdots\succ b_K\) of the arms,
unknown to the learner, such that \(\epsilon_{b_i,b_j}>0\) whenever
\(i<j\). Moreover, \(\epsilon_{i,j}\in(-1/2,1/2)\) for all arms \(i,j\).
\end{assumption}

For the analysis, we label the arms so that arm \(i\) is \(b_i\); thus
arm \(1\) is the best arm and \(\epsilon_{i,j}>0\) whenever \(i<j\).
The algorithms do not use this labeling.

\begin{assumption}[Relaxed stochastic transitivity]
\label{asmp:rst}
There is a known \(\gamma\ge1\) such that
\(\max\{\epsilon_{i,j},\epsilon_{j,k}\}\le\gamma\,\epsilon_{i,k}\)
for every \(i<j<k\).
\end{assumption}

\begin{assumption}[Stochastic triangle inequality]
\label{asmp:sti}
\(\epsilon_{i,k}\le\epsilon_{i,j}+\epsilon_{j,k}\) for every \(i<j<k\).
\end{assumption}

\begin{assumption}[Strong transitivity and triangle inequality at the best arm]
\label{asmp:sst-best}
For every \(1<j<k\),
\(\max\{\epsilon_{1,j},\epsilon_{j,k}\}\le\epsilon_{1,k}
\le\epsilon_{1,j}+\epsilon_{j,k}\).
\end{assumption}
Assumption~\ref{asmp:sst-best} follows from
Assumptions~\ref{asmp:rst} and~\ref{asmp:sti} with \(\gamma=1\), but it
constrains only triples that contain the best arm.
We develop two algorithms based on the Beat-the-Mean (BTM) approach of
\citet{yuejoachims2011}.
The first requires Assumptions~\ref{asmp:strict-order}--\ref{asmp:sti} and
handles \emph{relaxed} stochastic transitivity through single-arm elimination.
The second requires Assumptions~\ref{asmp:strict-order}
and~\ref{asmp:sst-best}; it exploits strong stochastic transitivity at the
best arm to eliminate multiple arms simultaneously and improve the
dependence on \(K\).
Throughout, \(T\ge K\ge2\), \(\epsilon_i:=\epsilon_{1,i}=P_{1,i}-\tfrac12\),
and \(H_{\mathrm p}:=\sum_{i=2}^K\epsilon_i^{-1}\).

\textbf{Active-set scores and shared rounding.}
For an active set \(W\subseteq[K]\) and \(i\in W\), define
\begin{equation}
\label{eq:finite-active-score}
  p_i(W):=\frac1{|W|}\sum_{j\in W}P_{i,j}.
\end{equation}
A comparison with a uniform opponent from \(W\) has mean \(p_i(W)\).
Both algorithms estimate these scores and round their estimates
using a shared random grid.
For \(h>0\) and \(u\in[0,1)\), let
\begin{equation}
\label{eq:finite-rounding}
  Q_h(x,u):=
  h\left(u+\left\lfloor\frac{x}{h}-u\right\rfloor\right).
\end{equation}
Rounding changes \(x\) by less than \(h\).
For \(U\sim\operatorname{Unif}[0,1)\), two fixed values \(x,y\)
round differently with probability at most \(|x-y|/h\).
Combining this property with concentration gives the replicable
mean estimates used below, with \(h=\tau/2\) for accuracy scale
\(\tau\).
The opponent-sampling variables and rounding offsets are mutually independent and independent of the comparison-noise variables. Each offset is revealed only when used.
The executions share opponent randomness and offsets, while
their comparison randomness is independent.

\begin{remark}[\textbf{Comparison with reward bandits}]
\label{rem:finite-reward-comparison}
Eliminating an arm in a stochastic reward bandit leaves the reward
distributions of the surviving arms unchanged.
In our uniform-opponent construction, eliminating an opponent
changes \(p_i(W)\) even though preference matrix is stationary.
Consequently, concentration and replicability must be established
for estimates whose target distributions change with the active set.
\end{remark}

\subsection{Relaxed Stochastic Transitivity}
\label{sec:finite-relaxed}

\paragraph{Why single-arm elimination?}
Under relaxed transitivity, the best arm need not maximize
\(p_i(W)\).
For example, \(P_{1,2}=0.55\), \(P_{1,3}=0.80\), and
\(P_{2,3}=0.95\) satisfy
Assumptions~\ref{asmp:strict-order}--\ref{asmp:sti} with
\(\gamma=1.5\), but
\(p_2([3])-p_1([3])=1/60\).
Thus, eliminating every arm separated from the maximum score
can discard the best arm, even with exact scores.
The following BTM inequalities instead support elimination of
a minimum-score arm \citep{yuejoachims2011}.

\begin{lemma}[Separation from worst arm]
\label{lem:finite-worst-separation}
Let \(1\in W\), \(|W|\ge2\), and let \(k(W)\) be the
worst-ranked arm in \(W\).
Write \(\eta(W):=\epsilon_{k(W)}\).
Under Assumptions~\ref{asmp:strict-order}--\ref{asmp:sti},
\begin{equation}
\label{eq:finite-worst-separation}
\begin{aligned}
  p_1(W)-p_{k(W)}(W)&\ge\eta(W),\\
  \max_{j\in W}\{p_j(W)-p_{k(W)}(W)\}
  &\le2\gamma^2\eta(W).
\end{aligned}
\end{equation}
\end{lemma}

Suppose every estimate has error at most \(\tau\).
If \(\tau<\eta(W)/2\), the best arm has a larger estimated
score than \(k(W)\), so it cannot be an estimated minimizer.
Otherwise, \eqref{eq:finite-worst-separation} implies
\(\max_j\widehat p_j-\widehat p_1\le4\gamma^2\tau\).
In either case, removing only an estimated minimum when the
estimated range exceeds \(6\gamma^2\tau\) preserves the best arm.
After removal, the scores are estimated for the new set.

\textbf{Algorithm and parameters.}
We call this procedure $\RepBTM$.
It stores an ordered list \(\mathcal S_i\) of opponent labels
and observations for each active leader \(i\).
Removing an arm deletes its list and every record involving it
as an opponent; all other records are retained.
For each leader \(i\), the shared seed supplies an independent
stream \((J_{i,s})_{s\ge1}\) of uniform labels in \([K]\).
Its pointer advances until the next label in the current \(W\),
skipping excluded labels without playing comparisons.
This produces uniform opponents and supports the retained-prefix
representation used in the analysis.

Set \(L:=1+\lceil\log_2T\rceil\),
\(\mathcal Q:=\{2^g:0\le g<L\}\), and
\(M:=K(K+1)L/2\).
For \(\rho\in[1/(2T),1)\), define
\begin{equation}
\label{eq:finite-relaxed-parameters}
\begin{aligned}
  \rho_{\mathrm R}&:=\frac\rho M,
  &\delta_{\mathrm R}&:=\frac1{4MT},\\
  \Lambda_{\mathrm R}
  &:=\frac{32\log(1/\delta_{\mathrm R})}
  {(\rho_{\mathrm R}-\delta_{\mathrm R})^2},
  &\tau(q)&:=\sqrt{\frac{\Lambda_{\mathrm R}}q}.
\end{aligned}
\end{equation}
Each interval with a fixed active set is a \emph{stage}, indexed
by \(\ell\).
The algorithm checks dyadic retained-sample counts in increasing
order, using a fresh shared offset \(U_{\ell,q,i}\) for every
stage--level--arm triple.
At most \(L\sum_{m=2}^Km\le M\) scalar estimates are computed.

\begin{algorithm}[t]
\caption{$\RepBTM$: single-arm elimination}
\label{alg:finite-relaxed}
\small
\begin{algorithmic}[1]
\REQUIRE \(K,T,\gamma,\rho\) and
\eqref{eq:finite-relaxed-parameters}.
\STATE \(W\gets[K]\), \(t\gets0\), \(\ell\gets1\),
\(\mathcal C\gets\varnothing\); initialize empty lists and
opponent-stream pointers.
\WHILE{\(t<T\) and \(|W|>1\)}
  \STATE \(\mathcal D\gets
  \{q\in\mathcal Q\setminus\mathcal C:
  q\le\min_{i\in W}|\mathcal S_i|\}\).
  \IF{\(\mathcal D\ne\varnothing\)}
    \STATE \(q\gets\min\mathcal D\).
    \STATE For every \(i\in W\), average its first \(q\)
    retained observations to obtain \(\overline X_i\), and set
    \(\widehat p_i\gets
    Q_{\tau(q)/2}(\overline X_i,U_{\ell,q,i})\).
    \STATE \(a\gets\arg\min_{i\in W}\widehat p_i\).
    \IF{\(\max_{i\in W}\widehat p_i-\widehat p_a
    >6\gamma^2\tau(q)\)}
      \STATE Remove \(a\) and all records involving \(a\);
      preserve the order of surviving records.
      \STATE \(\ell\gets\ell+1\);
      \(\mathcal C\gets\varnothing\).
    \ELSE
      \STATE \(\mathcal C\gets\mathcal C\cup\{q\}\).
    \ENDIF
  \ELSE
    \STATE \(i\gets\arg\min_{i\in W}|\mathcal S_i|\);
    take the next opponent \(j\in W\) from its stream.
    \STATE Compare \((i,j)\); append its opponent label and
    outcome to \(\mathcal S_i\), using \(1/2\) if \(i=j\);
    \(t\gets t+1\).
  \ENDIF
\ENDWHILE
\STATE If \(W=\{i\}\), compare \((i,i)\) until time \(T\).
\end{algorithmic}
\end{algorithm}

\begin{theorem}[Relaxed-transitivity guarantee]
\label{thm:finite-relaxed}
Suppose Assumptions~\ref{asmp:strict-order}--\ref{asmp:sti} hold.
For \(T\ge K\ge2\) and \(\rho\in[1/(2T),1)\),
Algorithm~\ref{alg:finite-relaxed} is \(\rho\)-replicable.
For a universal constant \(C>0\), with probability at least
\(1-1/(2T)\), it never eliminates \(b_1\) and
\begin{equation}
\label{eq:finite-relaxed-regret}
\begin{aligned}
  \mathbb E[R_T]
  &\le C\frac{\gamma^8M^2}{\rho^2}
  \log(4MT)H_{\mathrm p}+1=\widetilde{\mathcal O}\!\left(
  \frac{\gamma^8K^4H_{\mathrm p}}{\rho^2}\right).
\end{aligned}
\end{equation}
\end{theorem}
\begin{proof}[Proof sketch]
Full details are in Appendix~\ref{app:finite:proofs}.

\textbf{Step 1: Accuracy of retained estimates.} For each fixed set \(W\) and leader \(i\in W\), filter the leader's label stream to \(W\). Whenever the active set is \(W\), the retained list \(\mathcal S_i\) is a prefix of this filtered sequence, whose labels are i.i.d.\ uniform on \(W\) (Lemma~\ref{app:finite:prefix}); this avoids conditioning on the data-dependent active set. Hoeffding's inequality and a union bound over all \(K2^{K-1}L\) triples \((W,i,q)\) show that, with probability at least \(1-1/(4T)\), every realized estimate at level \(q\) is within \(\tau(q)\) of its active-set score. The union over exponentially many sets is affordable because \(\Lambda_{\mathrm R}\ge32M^2\log(4MT)\) (Lemma~\ref{app:finite:relaxed-accuracy}). On this event, Lemma~\ref{lem:finite-worst-separation} implies that the elimination rule never removes the best arm (Lemma~\ref{app:finite:relaxed-preservation}).

\textbf{Step 2: Bounded record counts.} On the same event, a check in \(W\) at any level \(q\ge B(W):=\lceil144\gamma^4\Lambda_{\mathrm R}/\eta(W)^2\rceil\) eliminates an arm, so during a stage with active set \(V\), sampling never raises a leader's count above \(2B(V)\). Since \(\eta(W)\le\gamma\eta(V)\) whenever \(1\in W\subseteq V\), every leader in a later set \(W\) holds at most \(d(W)=\mathcal O(\gamma^6\Lambda_{\mathrm R}/\eta(W)^2)\) records (Lemma~\ref{app:finite:counts}). A concentration bound on opponent-label counts, uniform over fixed sets, shows that with probability at least \(1-1/(4T)\) every elimination from \(W\) deletes at most \(4d(W)\) records (Lemma~\ref{app:finite:deleted-records}).

\textbf{Step 3: Regret.} Charge each comparison to the elimination that deletes its record, or to the terminal active set if its record is never deleted. Both endpoints of a record deleted from \(W\) lie in \(W\), so its comparison has regret at most \(\gamma\eta(W)\). If \(|W|=m\), then \(k(W)\ge m\) and hence \(\epsilon_m\le\gamma\eta(W)\), so the elimination from \(W\) is charged \(\mathcal O(\gamma^8\Lambda_{\mathrm R}/\epsilon_m)\). Each elimination removes one arm, so the pre-elimination sizes \(m\in\{2,\ldots,K\}\) are distinct, and summing gives \(\mathcal O(\gamma^8\Lambda_{\mathrm R}H_{\mathrm p})\); the terminal set is bounded in the same way. Hence \(R_T\le C\gamma^8\Lambda_{\mathrm R}H_{\mathrm p}\) with probability at least \(1-1/(2T)\), and \(\Lambda_{\mathrm R}\le128M^2\log(4MT)/\rho^2\) gives \eqref{eq:finite-relaxed-regret}.

\textbf{Step 4: Replicability.} Consider an execution that replaces each estimate at stage \(\ell\), level \(q\), and arm \(i\) by the rounded population score \(Q_{\tau(q)/2}(p_i(W),U_{\ell,q,i})\). Its sequence of active sets depends only on rounding offsets from earlier stages, so it is independent of opponent labels and comparison outcomes. For each (at most \(M\)) resulting triples, the shared-grid estimator makes the estimates of both executions equal the reference value with probability at least \(1-\rho_{\mathrm R}\) (Lemma~\ref{app:finite:mean}), so all of them agree with probability at least \(1-M\rho_{\mathrm R}=1-\rho\) (Lemma~\ref{app:finite:reference-agreement}). On this event, Lemma~\ref{app:finite:prefix} identifies each realized estimate with a reference prefix estimate, and induction over the algorithm's operations shows that both executions choose the same leaders, opponents, eliminations, and deleted records, and hence the same ordered pair in every round.
\end{proof}

\begin{remark}[Relation to standard Beat-the-Mean]
\label{rem:finite-btm-comparison}
The elimination rule and deletion follow BTM \citep{yuejoachims2011}.
Our analysis additionally controls adaptively retained estimates
and synchronizes the complete sampling and deletion process
across independent realizations.
\end{remark}

\subsection{Sharper Guarantees under Strong Stochastic Transitivity}
\label{sec:finite-sst}

Without replicability, Beat-the-Mean attains regret \(\mathcal O(\gamma^7K\log T/\epsilon_{\min})\), where \(\epsilon_{\min}:=\min_{i>1}\epsilon_i\) \citep{yuejoachims2011}. This is linear in \(K\), as are the regret bounds of standard algorithms for \(K\)-armed reward bandits \citep{auer2002finite}. With replicability, this parity breaks. For reward bandits, replicable elimination inflates the regret by a factor of \(K^2/\rho^2\) \citep{esfandiari2023replicable}, and this \(K^2/\rho^2\) overhead is also present in subsequent algorithms \citep{komiyama2024replicability,deb2026replicable}. In contrast, Theorem~\ref{thm:finite-relaxed} gives \(\widetilde{\mathcal O}(\gamma^8K^4H_{\mathrm p}/\rho^2)\), an overhead of \(K^4/\rho^2\) that remains even when \(\gamma=1\). Thus, although the non-replicable dueling and reward bounds have the same dependence on \(K\), the replicable dueling bound in Theorem~\ref{thm:finite-relaxed} is worse by a factor of \(K^2\) compared to \cite{komiyama2024replicability,deb2026replicable}.

This raises a natural question: \emph{can a replicable dueling bandit algorithm match the \(K^2/\rho^2\) overhead of replicable reward bandits?} We answer this affirmatively under strong stochastic transitivity, \(\gamma=1\). Algorithm~\ref{alg:finite-sst} below attains regret \(\widetilde{\mathcal O}(K^2H_{\mathrm p}/\rho^2)\) (Theorem~\ref{thm:finite-sst}). It changes both the elimination rule and the sampling schedule. Strong transitivity makes the best arm maximize the active-set score in every active set that contains it, which permits simultaneous elimination with fresh samples and reduces the number of rounded estimates from \(\mathcal O(K^2\log T)\) to \(\mathcal O(K\log T)\). 
\vspace{1ex}

\begin{lemma}[Score gaps under strong transitivity]
\label{lem:finite-sst-scores}
For \(1\in W\), define
\(d_i(W):=p_1(W)-p_i(W)\) and
\(r_i(W):=|\{j\in W:b_j\succeq b_i\}|\).
Then, for every \(i\in W\),
\begin{equation}
\label{eq:finite-sst-scores}
  \frac{r_i(W)}{|W|}\epsilon_i
  \le d_i(W)\le2\epsilon_i.
\end{equation}
Moreover, \(b_i\succ b_j\) implies
\(\epsilon_i\le\epsilon_j\).
\end{lemma}

In particular, the best arm maximizes \(p_i(W)\) in every
active set containing it.
If all estimates are \(\tau\)-accurate, its estimated score is
within \(2\tau\) of the estimated maximum.
It therefore survives simultaneous removal of arms whose
estimated deficits are at least \(4\tau\).
Every surviving arm has population score gap below \(6\tau\).
The rank factor in \eqref{eq:finite-sst-scores} will also control
the regret accumulated before elimination.

\textbf{Algorithm and parameters.}
The second algorithm, $\RepBTMSST$, uses phases
\(g=0,1,\ldots\), with \(n_g:=2^g\) fresh observations per
active leader.
Within a phase, the active set is fixed, leaders cycle in fixed
order, and opponents are independent uniform draws from that set.
At its checkpoint, all scores are rounded and elimination is
simultaneous.
Records are then discarded, and the phase index increases even
when arms are eliminated.

Fix \(\rho\in(0,1)\) and \(0<\delta<\rho/3\).
Set \(H:=1+\lfloor\log_2T\rfloor\), \(N:=KH\),
\(\rho_{\mathrm S}:=\rho/N\), and
\(\delta_{\mathrm S}:=\delta/N\).
Define
\begin{equation}
\label{eq:finite-sst-parameters}
  \Lambda_{\mathrm S}:=
  \frac{8\log(4/\delta_{\mathrm S})}
  {(\rho_{\mathrm S}-\delta_{\mathrm S})^2},
  \qquad
  \tau_g:=\sqrt{\frac{\Lambda_{\mathrm S}}{n_g}}.
\end{equation}
There are at most \(H\) phases and \(N\) scalar
estimator calls.
Opponent uniforms and offsets \(U_{g,i}\) are shared,
mutually independent, and revealed only when used.

\begin{algorithm}[t]
\caption{$\RepBTMSST$: simultaneous elimination}
\label{alg:finite-sst}
\small
\begin{algorithmic}[1]
\REQUIRE \(K,T,\rho,\delta\) and
\eqref{eq:finite-sst-parameters}.
\STATE \(W_0\gets[K]\), \(g\gets0\), \(t\gets0\).
\WHILE{\(t<T\)}
  \IF{\(W_g=\{i\}\)}
    \STATE Compare \((i,i)\) until time \(T\);
    \textbf{return}.
  \ENDIF
  \STATE \(n_g\gets2^g\),
  \(\tau_g\gets\sqrt{\Lambda_{\mathrm S}/n_g}\);
  initialize empty phase records.
  \FOR{\(s=1,\ldots,|W_g|n_g\)}
    \STATE If \(t=T\), \textbf{return}.
    \STATE Select the next cyclic leader \(i\);
    draw a fresh shared uniform opponent \(j\in W_g\).
    \STATE Compare \((i,j)\), record its outcome for \(i\),
    using \(1/2\) if \(i=j\), and set \(t\gets t+1\).
  \ENDFOR
  \STATE For every \(i\in W_g\), compute its phase mean
  \(\overline X_{i,g}\) and set
  \(\widehat p_{i,g}\gets
  Q_{\tau_g/2}(\overline X_{i,g},U_{g,i})\).
  \STATE \(W_{g+1}\gets
  \{i\in W_g:
  \max_{j\in W_g}\widehat p_{j,g}-\widehat p_{i,g}
  <4\tau_g\}\).
  \STATE Discard phase records; \(g\gets g+1\).
\ENDWHILE
\end{algorithmic}
\end{algorithm}

\begin{theorem}[Strong-transitivity guarantee]
\label{thm:finite-sst}
Suppose Assumptions~\ref{asmp:strict-order} and~\ref{asmp:sst-best}
hold.
For every \(T\ge K\ge2\), \(\rho\in(0,1)\), and
\(0<\delta<\rho/3\), Algorithm~\ref{alg:finite-sst} is
\(\rho\)-replicable.
With probability at least \(1-2\delta\), it never eliminates
\(b_1\) and \(R_T\le C\Lambda_{\mathrm S}H_{\mathrm p}\)
for a universal constant \(C>0\).
Taking \(\delta=\min\{\rho/4,1/T\}\) gives

\begin{equation}
\label{eq:finite-sst-regret}
  \mathbb E[R_T]
  =\widetilde{\mathcal O}\!\left(
  \frac{K^2H_{\mathrm p}}{\rho^2}\right).
\end{equation}
\end{theorem}
\begin{proof}[Proof sketch]
Full details are in Appendix~\ref{app:finite:sst}.

\textbf{Step 1: Accuracy with fresh samples.} Conditional on the history before phase \(g\), the active set \(W_g\) is fixed, opponents are i.i.d.\ uniform on \(W_g\), and every observation entering the phase-\(g\) estimates is collected after this history. The shared-grid estimator therefore makes each rounded estimate \(\tau_g\)-accurate with probability at least \(1-\delta_{\mathrm S}\) (Lemma~\ref{app:finite:mean}). Since there are at most \(N\) estimator calls (Lemma~\ref{app:finite:sst-schedule}), a union bound shows that all estimates are \(\tau_g\)-accurate on an event \(\mathcal E_{\mathrm S}\) of probability at least \(1-\delta\) (Lemma~\ref{app:finite:sst-accuracy}).

\textbf{Step 2: Survival of the best arm and the survivor bound.} On \(\mathcal E_{\mathrm S}\), Lemma~\ref{lem:finite-sst-scores} implies that the best arm maximizes \(p_i(W_g)\), so its estimate is within \(2\tau_g\) of the maximum estimate and it is never eliminated (Lemma~\ref{app:finite:sst-safety}). Every survivor of phase \(g\) satisfies \(d_i(W_g)<6\tau_g\) (Lemma~\ref{app:finite:sst-elimination}). Let \(m_g:=|W_g|\) and \(\eta_g:=\max_{i\in W_g}\epsilon_i\). The worst arm in \(W_{g+1}\) has rank at least \(m_{g+1}\) in \(W_g\), so the lower bound in 
\eqref{eq:finite-sst-scores} gives 
\begin{equation}
\label{eq:finite-survivor-bound}\eta_{g+1}<6\frac{m_g}{m_{g+1}}\tau_g
\end{equation} 
(Lemma~\ref{app:finite:sst-survivor-gap}). Thus a gap that is large relative to \(\tau_g\) can survive phase \(g\) only if the active set shrinks proportionally.

\textbf{Step 3: Exploration cost.} Let \(G\) be the last started phase and let \(c_g\) be the number of suboptimal arms whose last started phase is \(g\). Since \(\sum_{h\le g}n_h<2n_g\), each arm's total budget is at most twice its final-phase budget. With \(u_g:=\eta_g/\tau_g\) and \(v_g:=c_g\sqrt{n_g}\), this gives \(S\le2\sqrt{\Lambda_{\mathrm S}}\sum_gv_gu_g\) and \(\Lambda_{\mathrm S}H_{\mathrm p}\ge\sqrt{\Lambda_{\mathrm S}}\sum_gv_g/u_g\) (Lemma~\ref{app:finite:sst-cost-benchmark}). Summing \eqref{eq:finite-survivor-bound} over phases yields \(\sum_gv_gu_g\le43\sum_gv_g\) (Lemma~\ref{app:finite:sst-summed-barrier}), and Cauchy--Schwarz combines these bounds into 
\begin{equation}
\label{eq:finite-exploration-accounting}
S:=\sum_{g=0}^{G}n_g\sum_{i\in W_g}\epsilon_i\le3698\Lambda_{\mathrm S}H_{\mathrm p}
\end{equation} 
on \(\mathcal E_{\mathrm S}\), including a truncated final phase (Lemma~\ref{app:finite:sst-exploration-cost}).

\textbf{Step 4: Regret.} On \(\mathcal E_{\mathrm S}\), any exploitation rounds compare the best arm with itself and incur no regret. Each arm leads at most \(n_g\) times in phase \(g\), so the leaders' total gap is at most \(S\) (Lemma~\ref{app:finite:sst-leader}). Opponents are uniform on \(W_g\), so their expected total gap is also at most \(S\), and an exponential supermartingale bounds their realized total gap by \(2S+\log(1/\delta)\) with probability at least \(1-\delta\) (Lemma~\ref{app:finite:sst-opponent}). Hence \(R_T\le\tfrac32S+\tfrac12\log(1/\delta)\le6000\Lambda_{\mathrm S}H_{\mathrm p}\) with probability at least \(1-2\delta\). Taking \(\delta=\min\{\rho/4,1/T\}\) gives \(\Lambda_{\mathrm S}=\mathcal O(N^2\log(NT/\rho)/\rho^2)\) with \(N=\mathcal O(K\log T)\), which proves \eqref{eq:finite-sst-regret}.

\textbf{Step 5: Replicability.} If both executions start phase \(g\) with the same active set and start time, the cyclic leader schedule and the shared opponent uniforms produce identical pairs throughout the phase, regardless of the outcomes. If their rounded scores also agree at the checkpoint, the elimination rule produces the same next active set. Hence the trajectories can differ only if the rounded vectors first disagree at some completed checkpoint. Conditional on the common pre-phase history and opponent labels, the observations of phase \(g\) are fresh and independent across executions, so the shared-grid estimator bounds the disagreement probability of each coordinate by \(\rho_{\mathrm S}\) (Lemma~\ref{app:finite:mean}). A union bound over at most \(K\) arms and \(H\) phases gives failure probability at most \(HK\rho_{\mathrm S}=\rho\) (Lemma~\ref{app:finite:sst-phase-coupling}).
\end{proof}

\begin{remark}[\textbf{Source of the improvement}]
\label{rem:finite-improvement}
The improvement combines the global phase schedule with
\eqref{eq:finite-exploration-accounting}.
The number of scalar estimates decreases from
\(M=\mathcal O(K^2\log T)\) to
\(N=\mathcal O(K\log T)\).
Since the required sample size grows quadratically with the inverse
per-call replicability budget, this replaces a \(K^4/\rho^2\)
prefactor by \(K^2/\rho^2\).
The aggregate analysis is essential: using only
\(d_i(W)\ge\epsilon_i/|W|\) separately for each arm could
introduce another \(K^2\) sample penalty.
The rank-sensitive survivor bound avoids that loss.
Moreover, resetting samples costs only a constant factor relative
to each arm's final phase budget, since
\(\sum_{h=0}^g n_h<2n_g\).
The resulting guarantee has the same \(K^2/\rho^2\) factor
as the basic replicable elimination bound for reward bandits
\citep{esfandiari2023replicable}. Whether a similar \(K^2/\rho^2\) bound can be obtained for the relaxed $\gamma\geq 1$ case, is an open question.
\end{remark}

%% file: section/linear.tex
Our algorithm, $\RepMaxInP$, extends the replicable generalized linear estimation and determinant-triggered batching of \citet{deb2026replicable} to linear dueling bandits by combining it with the Maximum-Informative-Pair rule \citep{saha2021maxinp}. Preferences satisfy \(P_t(i,j)=\mu((x_{t,i}-x_{t,j})^\top\theta^\star)\). For known \(L,S>0\), assume \(\|x_{t,i}\|_2\le L\) and \(\theta^\star\in\Theta\), where \(\Theta\subset\mathbb R^d\) is a known compact convex set containing \(0\), with \(\sup_{\theta\in\Theta}\|\theta\|_2\le S\). The known increasing, continuously differentiable link \(\mu:\mathbb R\to(0,1)\) satisfies \(\mu(-u)=1-\mu(u)\) and \(\mu'(z^\top\theta)\ge c_\mu\) for every \(\theta\in\Theta\) and \(\|z\|_2\le2L\), with known \(c_\mu\in(0,1]\). The logistic link is one example. Writing \(x_t=x_{t,i_t}\), \(y_t=x_{t,j_t}\), and \(x_t^\star\) for a utility-maximizing action, we measure performance by utility regret:
\begin{equation}
\label{eq:utility-regret}
\begin{aligned}
R_T^{\mathrm{util}}&:=\sum_{t=1}^T r_t^{\mathrm{util}},\\
r_t^{\mathrm{util}}&:=\Big(x_t^\star-\frac{x_t+y_t}{2}\Big)^\top\theta^\star.
\end{aligned}
\end{equation}

\textbf{Algorithm.}
Our algorithm, $\RepMaxInP$, combines the Maximum-Informative-Pair rule \citep{saha2021maxinp} with batched replicable estimation. Let \(z_t=x_t-y_t\) and \(V_t=\lambda I_d+\sum_{s<t}z_sz_s^\top\), where \(\lambda>0\). At batch start \(\tau_b\), it fits a generalized linear estimate using past comparisons, applies a fresh shared randomly shifted grid in the coordinates induced by \(V_{\tau_b}^{1/2}\), and maps back to obtain \(\widetilde\theta_b\). The resulting confidence radius is \(\Gamma_b=\beta_{\tau_b}(\delta/B)(1+dB/(\rho-2\delta))\), where \(\beta_t\) is the statistical radius and \(B=\widetilde{\mathcal O}(d)\) bounds the number of batches. The exact details and proof appear in Appendix~\ref{app:repmaxinp}.

Within the batch, the estimate, radius, and selection matrix remain fixed. The promising indices are
\begin{equation}
\label{eq:repmaxinp-promising}
\begin{aligned}
&\mathcal C_t:=\bigl\{i\in[K_t]:\ \forall j\in[K_t],\\
&\quad(x_{t,i}-x_{t,j})^\top\widetilde\theta_b+\Gamma_b\|x_{t,i}-x_{t,j}\|_{V_{\tau_b}^{-1}}\ge0\bigr\}.
\end{aligned}
\end{equation}
The learner selects the pair in \(\mathcal C_t^2\) maximizing \(\|x_{t,i}-x_{t,j}\|_{V_{\tau_b}^{-1}}\), with tie-breaking by index. A new batch starts whenever \(t<T\) and \(\det(V_{t+1})>e\det(V_{\tau_b})\).

\begin{theorem}[Linear dueling guarantee]
\label{thm:repmaxinp}
Under the model above, for every \(\rho\in(0,1)\) and \(0<\delta<\rho/3\), \textnormal{RepMaxInP} is \(\rho\)-replicable and uses at most \(B\) batches. For fixed \(L,S,\lambda\), with probability at least \(1-\delta\), every utility-maximizing index remains promising and

\begin{equation}
\label{eq:repmaxinp-rate}
R_T^{\mathrm{util}}=\widetilde{\mathcal O}\!\left(\frac{d^3}{c_\mu\rho}\sqrt T\right).
\end{equation}
Here \(\widetilde{\mathcal O}\) suppresses logarithmic factors, including those in \(1/\delta\). Choosing \(\delta=\min\{\rho/4,1/T\}\) gives the same bound on \(\mathbb E[R_T^{\mathrm{util}}]\). Neither guarantee has explicit dependence on the numbers of available actions \(K_t\).
\end{theorem}


%% file: section/experiments.tex
\begin{figure*}[t]
  \centering
  \includegraphics[width=\textwidth]{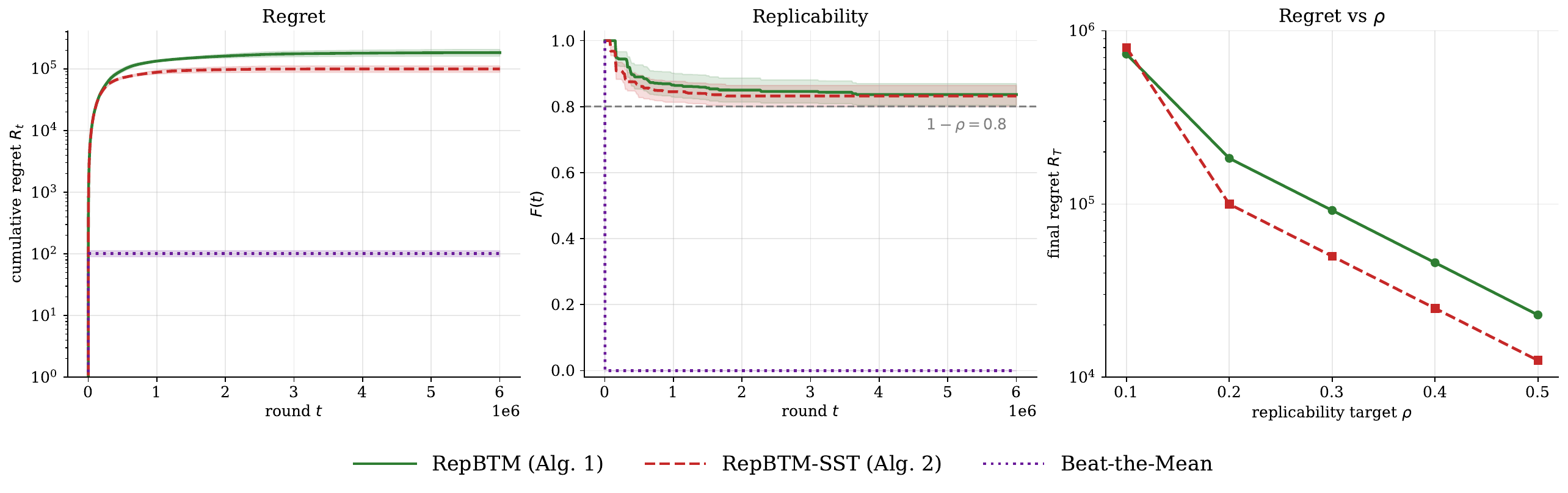}
  \caption{Left: cumulative regret at $\rho=0.2$ over the first $6\times10^{6}$ rounds; the curves stay flat up to $T$. Middle: fraction $F(t)$ of paired executions with identical trajectories up to round $t$, with the requirement $1-\rho$ dashed. Right: final regret against $\rho$. Bands and bars are $95\%$ intervals across $30$ test instances.}
  \label{fig:experiments}
\end{figure*}

This section provides supplementary numerical illustrations of our proposed approach. We examine how shared rounding affects agreement between paired executions and the regret incurred before commitment. The experiments use tuned confidence widths and, for the language-model study, the implementation variant specified in Appendix~\ref{app:experiments}.

\subsection{Synthetic Preferences}
\label{sec:synthetic-exp}

We compare RepBTM (Algorithm~1) and RepBTM-SST (Algorithm~2) with the non-replicable BTM~\citep{yuejoachims2011} on Bradley--Terry--Luce instances with $K=5$ arms. We run pairs of executions that share internal randomness but observe independent comparison outcomes, and report the fraction $F(t)$ of pairs whose trajectories coincide up to round $t$. Confidence widths are tuned on validation instances; details are in Appendix~\ref{app:experiments}.

Figure 1 illustrates the trade-off between trajectory agreement and exploration cost. At $\rho=0.2$, both replicable methods achieve aggregate empirical agreement above the target, whereas no paired BTM execution agrees throughout the horizon. Their cumulative regret curves plateau after commitment to the best arm, with $\RepBTMSST$ achieving lower regret than $\RepBTM$.
\subsection{Preferences from a Language Model}
\label{sec:llm-experiment}

\begin{figure*}[t]
  \centering
  \includegraphics[width=\textwidth]{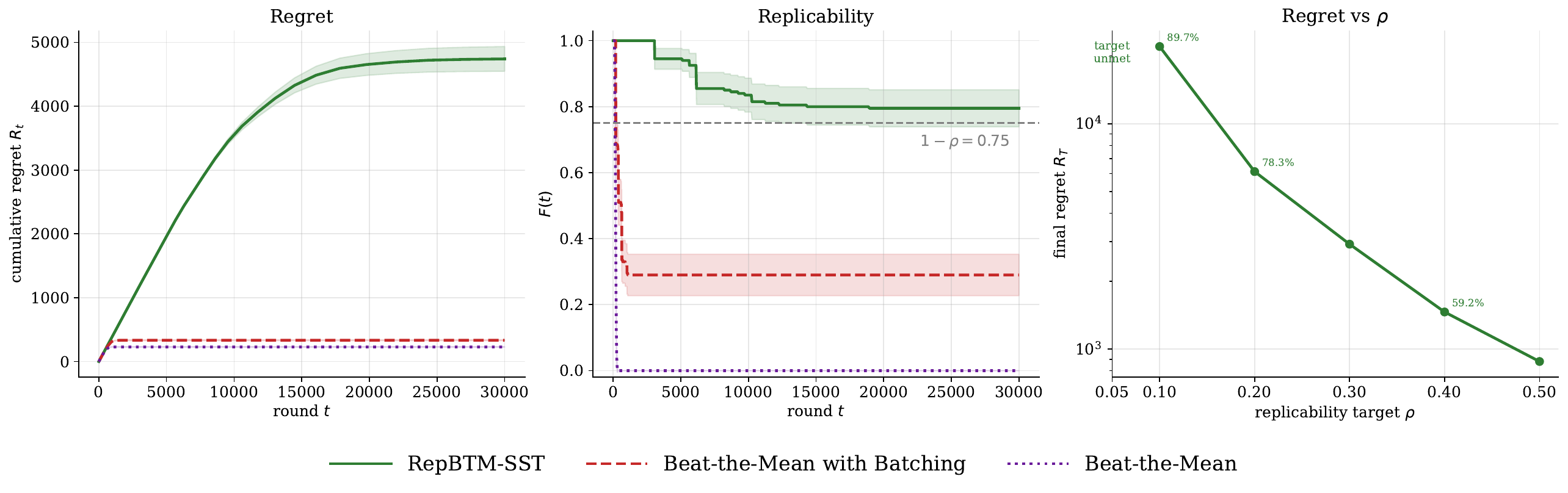}
  \caption{Left: cumulative regret at $\rho=0.25$ over $T=3\times10^{4}$ rounds. Middle: fraction $F(t)$ of paired executions with identical trajectories up to round $t$, with the requirement $1-\rho$ dashed. Right: final regret at $T=3\times10^{5}$ against $\rho$; labels give $F(T)$ where it is below $1-\rho$. Bands are $95\%$ intervals across paired runs.}
  \label{fig:llm}
\end{figure*}

We also evaluate $\RepBTMSST$ on preferences elicited from a language model. First, Qwen3.8-27B is used to summarize Wikipedia articles for three audiences (a five-year-old, a humanities student and a domain expert) at two targets lengths (short and long), thus forming $K=6$ arms. In each round $t$, acting as a humanities student who prefers long summaries, a different instance of the same model is then used as a judge to poll for pairwise feedback given two summaries of the same article. The regret here is measured against the long humanities-student arm, that is, the ground truth that also happens to be the Condorcet winner. Finally, as an ablation exercise, we also run a version of $\RepBTMSST$ without randomized rounding, which provides non-trivial but insufficient replicability, thus reinforcing the necessity of both batching and rounding. 

At $\rho=0.25$, Figure~\ref{fig:llm} shows empirical full-trajectory agreement of $0.795$ for $\RepBTMSST$, compared with $0.290$ for the unrounded batched baseline and $0$ for BTM. The regret curves plateau after commitment to the winner. The right panel illustrates how the final regret decreases as the replicability requirement is relaxed by increasing $\rho$. Details are in Appendix~\ref{app:llm}.

%% file: section/conclusion.tex
We established theoretical guarantees for trajectory replicability in stochastic dueling bandits. Our results show that agreement of the entire comparison sequence is compatible with sublinear regret for finite arm, and linear preference models.
We designed \(\RepBTM\), which achieves regret \(\widetilde{\mathcal O}(\gamma^8K^4H_{\mathrm p}/\rho^2)\) under relaxed stochastic transitivity, and \(\RepBTMSST\), which achieves \(\widetilde{\mathcal O}(K^2H_{\mathrm p}/\rho^2)\) under strong stochastic transitivity. For linear dueling bandits, we designed RepMaxInP, which achieves utility regret \(\widetilde{\mathcal O}(d^3\sqrt{T}/(c_\mu\rho))\).

Several questions remain open. Under relaxed transitivity, the best arm need not maximize the active-set score, so single-arm elimination requires \(\mathcal O(K^2\log T)\) rounded estimates and incurs a \(K^4/\rho^2\) overhead; whether a different elimination rule or score achieves \(K^2/\rho^2\) for \(\gamma>1\) is open. More fundamentally, it is unknown whether the \(K^2/\rho^2\) overhead is necessary, even for \(\gamma=1\) or for reward bandits: to our knowledge, no lower bound for \(\rho\)-replicable bandits depends on \(K\), and the existing lower bound is proved for two arms \citep{komiyama2024replicability}. It is also open whether the overhead can be separated from the leading \(\log T\) term in dueling bandits, as \citet{komiyama2024replicability} show for reward bandits. In the linear setting, improving the \(d^3\) dependence or proving matching lower bounds would clarify the price of replicability under preference feedback. Neural dueling bandits \citep{verma2025neural} model nonlinear utilities with neural networks, building on neural contextual bandits \citep{zhou2020neural,zhang2021neural,deb2024contextual}; making them replicable would require replicable neural estimators and confidence bounds rather than a replicable finite-dimensional parameter. \citet{bollini2026replicable} recently initiated replicability in constrained bandits, and extending replicable dueling algorithms to constrained, conservative, and safe settings, where comparisons must also satisfy cost or safety requirements \citep{linCBwK,conservative_linear_bandits,Moradipari2019SafeLT,deb2025conservative,deb2025thompson,deb24a}, is a natural direction. Finally, natural extensions include weaker notions of a winner that do not require a strict preference order and replicable active collection of preference data for training language models \citep{das2025active}.

%% file: appendix/repbtm.tex
We prove Lemmas~\ref{lem:finite-worst-separation} and~\ref{lem:finite-sst-scores} and Theorems~\ref{thm:finite-relaxed} and~\ref{thm:finite-sst}. Throughout, \(\epsilon_i=P_{1,i}-1/2\), \(\epsilon_1=0\), and \(H_{\mathrm p}=\sum_{i=2}^K\epsilon_i^{-1}\). The strict preference order gives \(0<\epsilon_i<1/2\) for every \(i>1\). All logarithms without a subscript are natural. The two algorithms use exactly the sampling, rounding, and elimination rules in Section~\ref{sec:finite-dueling}.

\subsection{Shared-grid estimation}
\label{app:finite:rounding}

We first establish the estimation facts used by both algorithms. Recall the rounding map in \eqref{eq:finite-rounding}:
\[
  Q_h(x,u) =h\left(u+\left\lfloor\frac{x}{h}-u\right\rfloor\right).
\]

\begin{lemma}[Rounding error and agreement]
\label{app:finite:grid}
For every \(h>0\), \(x\in\mathbb R\), and \(u\in[0,1)\), \(0\le x-Q_h(x,u)<h\). For fixed \(x,y\in\mathbb R\) and \(U\sim\operatorname{Unif}[0,1)\),
\begin{equation}
\label{app:finite:grid-agreement}
  \Pr\bigl(Q_h(x,U)\ne Q_h(y,U)\bigr) \le\min\{1,|x-y|/h\}.
\end{equation}
More generally, a fixed interval of length \(a\) contains a grid boundary with probability at most \(\min\{1,a/h\}\).
\end{lemma}

\begin{proof}
Writing \(k=\lfloor x/h-u\rfloor\) gives \(h(u+k)\le x<h(u+k)+h\), proving the first assertion. For \(x\le y\), the rounded values can differ only if a boundary \(h(U+k)\), \(k\in\mathbb Z\), belongs to \((x,y]\). A union bound and a change of variable give
\begin{align*}
  &\Pr\bigl(\exists k\in\mathbb Z: x<h(U+k)\le y\bigr)\\
  &\quad\le \sum_{k\in\mathbb Z}\int_0^1 \mathbf{1}\{x<h(u+k)\le y\}\,du\\
  &\quad= \int_{\mathbb R}\mathbf{1}\{x<hv\le y\}\,dv =\frac{y-x}{h}.
\end{align*}
The same argument applies to any fixed interval. Boundary endpoints have probability zero, so their inclusion does not change the bound. The case \(y<x\) follows by symmetry.
\end{proof}

\begin{lemma}[Accuracy and two forms of agreement]
\label{app:finite:mean}
Fix \(0<\beta<\alpha<1\), \(\tau>0\), and an integer \(n\ge1\) satisfying
\begin{equation}
\label{app:finite:mean-budget}
  n\tau^2\ge \frac{8\log(4/\beta)}{(\alpha-\beta)^2}.
\end{equation}
Let \(h=\tau/2\), and let \(U\) be uniform on \([0,1)\) and independent of all observations and opponent labels. Write \(\overline X^{(r)}=n^{-1}\sum_{s=1}^nX_s^{(r)}\) and \(\widehat p^{(r)}=Q_h(\overline X^{(r)},U)\). The following properties hold.
\begin{enumerate}
\item
If \(X_1^{(r)},\ldots,X_n^{(r)}\) are independent, take values in \([0,1]\), and have common mean \(\mu\), then
\[
  \Pr\bigl(|\widehat p^{(r)}-\mu|>\tau\bigr)\le\beta.
\]
\item
Fix an opponent sequence \(j_1,\ldots,j_n\) for leader \(i\). Suppose the observations, conditional on these labels, are independent across samples and executions, belong to \([0,1]\), and satisfy \(\mathbb E[X_s^{(r)}]=P_{i,j_s}\). Then
\[
  \Pr\bigl(\widehat p^{(1)}\ne\widehat p^{(2)} \mid J_1=j_1,\ldots,J_n=j_n\bigr)\le\alpha.
\]
\item
Suppose instead that the shared labels \(J_s\) are independent uniform draws from a fixed set \(W\) containing \(i\), and that, conditional on these labels, the observations obey the model in the second part. With \(\mu=p_i(W)\),
\begin{equation}
\label{app:finite:mean-reference}
  \Pr\bigl( \widehat p^{(1)}\ne Q_h(\mu,U) \text{ or } \widehat p^{(2)}\ne Q_h(\mu,U) \bigr)\le\alpha.
\end{equation}
\end{enumerate}
\end{lemma}

\begin{proof}
For the first assertion, Hoeffding's inequality gives
\[
  \Pr\bigl(|\overline X^{(r)}-\mu|>\tau/2\bigr) \le2e^{-n\tau^2/2} \le2(\beta/4)^4\le\beta.
\]
Here we used \(\alpha-\beta<1\) and \eqref{app:finite:mean-budget}. Rounding changes the empirical mean by less than \(\tau/2\), which proves the claim.

For the second assertion, condition on the labels and set \(\mu_J=n^{-1}\sum_{s=1}^nP_{i,j_s}\) and \(a=(\alpha-\beta)\tau/4\). The observations need not have identical means across \(s\). Hoeffding's inequality for each execution and a union bound yield
\begin{align*}
  &\Pr\bigl( \max_{r\in\{1,2\}}|\overline X^{(r)}-\mu_J|>a \mid J_1=j_1,\ldots,J_n=j_n\bigr)\\
  &\qquad\le4e^{-2na^2} \le4e^{-\log(4/\beta)}=\beta.
\end{align*}
On the complementary event, the two empirical means differ by at most \(2a\). Conditional on the observations, \(U\) is still uniform. Lemma~\ref{app:finite:grid} bounds its disagreement probability by \(2a/h=\alpha-\beta\). Adding the probability \(\beta\) of the exceptional event proves the second assertion.

For the third assertion, we first establish concentration around \(\mu\) after averaging over the labels. Fix an execution \(r\), write \(J=(J_1,\ldots,J_n)\), and let \(\lambda\in\mathbb R\). Convexity gives \(e^{\lambda x}\le1-x+xe^\lambda\) for \(x\in[0,1]\). Conditional independence of the observations and their prescribed conditional means therefore imply
\begin{align*}
  &\mathbb E\!\left[e^{\lambda\sum_{s=1}^n X_s^{(r)}}\mid J\right]\\
  &\qquad\le\prod_{s=1}^n\left(1-P_{i,J_s}+P_{i,J_s}e^\lambda\right).
\end{align*}
Because the labels are independent and uniform on \(W\), averaging this product over \(J\) yields
\[
  \mathbb E\!\left[e^{\lambda\sum_{s=1}^n X_s^{(r)}}\right]\le(1-\mu+\mu e^\lambda)^n.
\]
Hoeffding's lemma for a Bernoulli variable of mean \(\mu\) gives \(e^{-\lambda\mu}(1-\mu+\mu e^\lambda)\le e^{\lambda^2/8}\). Consequently,
\[
  \mathbb E\!\left[e^{\lambda\sum_{s=1}^n(X_s^{(r)}-\mu)}\right]\le e^{n\lambda^2/8}.
\]
Applying the exponential Markov inequality with \(\lambda=4a\) for the upper tail and \(\lambda=-4a\) for the lower tail gives \(\Pr(|\overline X^{(r)}-\mu|>a)\le2e^{-2na^2}\). A union bound over the two executions, which does not require independence between their empirical means, now gives
\[
  \Pr\bigl( \max_{r\in\{1,2\}}|\overline X^{(r)}-\mu|>a \bigr)\le\beta.
\]
Otherwise, the interval containing \(\mu\) and both empirical means has length at most \(2a\). Conditional on these means, the probability that this interval contains a grid boundary is at most \(2a/h=\alpha-\beta\). In the absence of such a boundary, all three rounded values coincide. This proves \eqref{app:finite:mean-reference}.
\end{proof}

The parameters of both algorithms satisfy \eqref{app:finite:mean-budget}. For Algorithm~\ref{alg:finite-sst}, this is exactly \(n_g\tau_g^2=\Lambda_{\mathrm S}\), with \((\alpha,\beta)=(\rho_{\mathrm S},\delta_{\mathrm S})\). For Algorithm~\ref{alg:finite-relaxed}, \(q\tau(q)^2=\Lambda_{\mathrm R}\), \(0<\delta_{\mathrm R}\le\rho_{\mathrm R}/2<1/2\), and
\[
  32\log(1/\delta_{\mathrm R}) \ge8\log(4/\delta_{\mathrm R}).
\]
Hence the lemma applies with \((\alpha,\beta)=(\rho_{\mathrm R},\delta_{\mathrm R})\).

\subsection{Relaxed stochastic transitivity}
\label{app:finite:relaxed}

This subsection uses Assumptions~\ref{asmp:strict-order}--\ref{asmp:sti} for every ordered triple, with \(\gamma\ge1\). For \(1\in W\) and \(|W|\ge2\), write \(k(W)=\max W\) and \(\eta(W)=\epsilon_{k(W)}\), using the preference-order indices only for analysis. The algorithm does not know this order.

\subsubsection{Preference inequalities and preservation of the best arm}

\begin{proof}[Proof of Lemma~\ref{lem:finite-worst-separation}]
Fix \(W\), write \(k=k(W)\), and set \(\eta=\epsilon_k\). For every \(h\in W\), the triangle inequality, or equality when \(h\in\{1,k\}\), gives
\[
  \epsilon_{1,h}+\epsilon_{h,k}\ge\epsilon_{1,k}=\eta.
\]
Averaging over \(h\) proves \(p_1(W)-p_k(W)\ge\eta\).

For the upper bound, fix \(j\in W\) and write
\[
  p_j(W)-p_k(W) =\frac1{|W|}\sum_{h\in W} (\epsilon_{j,h}+\epsilon_{h,k}).
\]
First, \(\epsilon_{h,k}\le\gamma\eta\) for every \(h\in W\). For \(1<h<k\) this follows from relaxed transitivity on \((1,h,k)\); the endpoints follow directly. If \(h\le j\), then \(\epsilon_{j,h}\le0\). If \(j<h\), relaxed transitivity and the endpoint cases give
\[
  \epsilon_{j,h}\le\gamma\epsilon_h \le\gamma^2\eta.
\]
Indeed, the first inequality follows from \((1,j,h)\) when \(j>1\), and the second from \((1,h,k)\) when \(h<k\). When \(j=1\) or \(h=k\), the corresponding inequality uses only \(\gamma\ge1\). Every summand is therefore at most \((\gamma^2+\gamma)\eta\le2\gamma^2\eta\). Averaging proves the upper bound.
\end{proof}

We will also use two consequences of relaxed transitivity:
\begin{equation}
\label{app:finite:nested-gaps}
\begin{aligned}
  \epsilon_i&\le\gamma\eta(W) &&(i\in W),\\
  \eta(W)&\le\gamma\eta(V) &&(1\in W\subseteq V,\ |W|\ge2).
\end{aligned}
\end{equation}
For the first inequality, apply the assumption to \((1,i,k(W))\) when its indices are distinct; the other cases are immediate. For the second, use \(k(W)\le k(V)\) and apply the same argument to the two worst-ranked arms. In particular, this argument does not assert that \(\eta(W)\) decreases when arms are removed.

\begin{lemma}[A correct check preserves the best arm]
\label{app:finite:relaxed-preservation}
Suppose \(1\in W\) and all estimates at a check satisfy \(|\widehat p_i-p_i(W)|\le\tau\). Removing an estimated minimum only when the estimated range exceeds \(6\gamma^2\tau\) does not eliminate arm \(1\).
\end{lemma}

\begin{proof}
Write \(k=k(W)\) and \(\eta=\eta(W)\). If \(\tau<\eta/2\), Lemma~\ref{lem:finite-worst-separation} implies
\[
  \widehat p_1-\widehat p_k \ge p_1(W)-p_k(W)-2\tau \ge\eta-2\tau>0.
\]
Thus arm \(1\) cannot be an estimated minimizer. If \(\tau\ge\eta/2\), then, for every \(j\in W\),
\begin{align*}
  \widehat p_j-\widehat p_1 &\le p_j(W)-p_1(W)+2\tau\\
  &\le(2\gamma^2-1)\eta+2\tau\\
  &\le4\gamma^2\tau<6\gamma^2\tau.
\end{align*}
If arm \(1\) were an estimated minimizer, this would make the estimated range smaller than the elimination threshold. The best arm therefore survives in both cases.
\end{proof}

\subsubsection{Retained prefixes and simultaneous accuracy}

We realize the sampling model on infinite arrays. For each leader \(i\), let \((J_{i,s})_{s\ge1}\) be its shared independent uniform labels in \([K]\). Take independent uniforms \(Z_{i,s}^{(r)}\), for \(r\in\{1,2\}\), independent of all labels and rounding offsets, and define the recorded values
\[
  X_{i,s}^{(r)}=
\begin{cases}
  1/2,&J_{i,s}=i,\\
  \mathbf{1}\{Z_{i,s}^{(r)}<P_{i,J_{i,s}}\},&J_{i,s}\ne i.
\end{cases}
\]
The actual self-comparison outcome can still be generated as a fresh Bernoulli variable; Algorithm~\ref{alg:finite-relaxed} records its mean \(1/2\) instead. Unused array entries are auxiliary variables. Revealing a fresh label and its value whenever a comparison is played has exactly the prescribed feedback distribution.

For a fixed nonempty set \(W\) and leader \(i\in W\), filter the infinite array to entries whose labels belong to \(W\), preserving their order. Write the resulting sequence as \((J_{i,W,s},X_{i,W,s}^{(r)})_{s\ge1}\) and let
\[
  \overline X_{i,W,q}^{(r)} =\frac1q\sum_{s=1}^qX_{i,W,s}^{(r)}.
\]
For this fixed set, the retained labels are independent uniform draws from \(W\). Within each execution, the filtered observations are independent and have mean \(p_i(W)\). Arrays for distinct leaders are independent before any data-dependent selection of an active set.

\begin{lemma}[Retained-prefix representation]
\label{app:finite:prefix}
Whenever an execution has active set \(W\), its retained list \(\mathcal S_i\), for each \(i\in W\), is a prefix of the sequence filtered to \(W\).
\end{lemma}

\begin{proof}
Fix such a leader \(i\), and consider any earlier sampling step with that leader and active set \(V\). Nestedness gives \(W\subseteq V\). A label skipped at that step lies outside \(V\), so it also lies outside \(W\). Every consumed label in \(W\) therefore produced a comparison. Both endpoints of that comparison survive in \(W\), so its record has not been deleted. Conversely, a retained record must have its opponent in \(W\). Thus the retained records are exactly the entries with labels in \(W\) among the consumed prefix of the original array. The deletion rule preserves their order.
\end{proof}

For one execution, define \(\mathcal E_{\mathrm R}\) by
\begin{equation}
\label{app:finite:uniform-prefix-event}
\begin{gathered}
  \left|\overline X_{i,W,q}-p_i(W)\right| \le\frac{\tau(q)}2,\\
  W\ne\varnothing,\quad i\in W,\quad q\in\mathcal Q.
\end{gathered}
\end{equation}
This event ranges over fixed sets, including sets never visited by the algorithm.

\begin{lemma}[Uniform retained-sample accuracy]
\label{app:finite:relaxed-accuracy}
\(\Pr(\mathcal E_{\mathrm R})\ge1-1/(4T)\). On \(\mathcal E_{\mathrm R}\), every realized rounded estimate at level \(q\) is within \(\tau(q)\) of its active-set score, and arm \(1\) is never eliminated.
\end{lemma}

\begin{proof}
For each fixed \((W,i,q)\), Hoeffding's inequality gives
\[
  \Pr\left( |\overline X_{i,W,q}-p_i(W)|>\tau(q)/2 \right)\le2e^{-\Lambda_{\mathrm R}/2}.
\]
There are \(K2^{K-1}L\) such triples. Since \(\rho_{\mathrm R}-\delta_{\mathrm R}<1/M\),
\begin{equation}
\label{app:finite:large-relaxed-budget}
  \Lambda_{\mathrm R}\ge32M^2\log(4MT).
\end{equation}
Using \(KL\le M\), \(K\le M\), and \(T\ge2\), we obtain
\begin{align*}
  \log(8TK2^{K-1}L) &\le\log(4MT)+K\log2\\
  &\le(K+1)\log(4MT) \le\Lambda_{\mathrm R}/2.
\end{align*}
Consequently,
\begin{equation}
\label{app:finite:uniform-prefix-probability}
  \Pr(\mathcal E_{\mathrm R}^c) \le2K2^{K-1}Le^{-\Lambda_{\mathrm R}/2} \le\frac1{4T}.
\end{equation}
Lemma~\ref{app:finite:prefix} identifies every realized mean with one of the prefixes in \eqref{app:finite:uniform-prefix-event}. Rounding contributes less than \(\tau(q)/2\). Finally, apply Lemma~\ref{app:finite:relaxed-preservation} inductively at all checks, starting from \(W=[K]\). This proves preservation of the best arm.
\end{proof}

\subsubsection{Retained counts and deleted records}

\begin{lemma}[A sufficient count for elimination]
\label{app:finite:threshold}
For \(1\in W\), \(|W|\ge2\), define
\[
  B(W)=\left\lceil \frac{144\gamma^4\Lambda_{\mathrm R}}{\eta(W)^2} \right\rceil, \qquad q_\star(W)=2^{\lceil\log_2 B(W)\rceil}.
\]
On \(\mathcal E_{\mathrm R}\), a check in \(W\) at any level \(q\ge B(W)\) eliminates an arm. Moreover, \(B(W)\le q_\star(W)<2B(W)\).
\end{lemma}

\begin{proof}
At such a check, \(\tau(q)\le\eta(W)/(12\gamma^2)\). Accuracy and Lemma~\ref{lem:finite-worst-separation} give
\begin{align*}
  \max_i\widehat p_i-\min_i\widehat p_i &\ge\widehat p_1-\widehat p_{k(W)}\\
  &\ge\eta(W)-2\tau(q)\\
  &\ge\tfrac56\eta(W) >\tfrac12\eta(W) \ge6\gamma^2\tau(q).
\end{align*}
The dyadic inequalities follow from the definition of \(q_\star(W)\).
\end{proof}

For the count analysis, set
\begin{equation}
\label{app:finite:count-scale}
\begin{gathered}
  A(W)=\frac{\gamma^6\Lambda_{\mathrm R}}{\eta(W)^2},\\
  d(W)=\lceil300A(W)\rceil, \qquad D(W)=4d(W).
\end{gathered}
\end{equation}
These are deterministic functions of a fixed set and the instance, used only in the analysis. In particular, \(A(W)>4\Lambda_{\mathrm R}>1\), so \(d(W)\le301A(W)\) and \(d(W)\ge\Lambda_{\mathrm R}\).

\begin{lemma}[Uniform bound on retained counts]
\label{app:finite:counts}
On \(\mathcal E_{\mathrm R}\), whenever the current set is \(W\) with \(|W|\ge2\), every active leader has at most \(d(W)\) records. This includes the moments immediately before elimination and at the horizon.
\end{lemma}

\begin{proof}
During a stage with set \(V\), sampling increments a leader with minimum retained count. If \(q_\star(V)\le T\), then \(q_\star(V)\in\mathcal Q\). The algorithm cannot sample while the minimum count is at least \(q_\star(V)\): it first examines all ready unchecked levels, and its check at \(q_\star(V)\) must eliminate an arm by Lemma~\ref{app:finite:threshold}. That level could not have been checked earlier without ending the stage. Every newly incremented count is therefore at most \(q_\star(V)\). If \(q_\star(V)>T\), the same upper bound holds because no leader can have more than \(T\) records.

An inherited count is unchanged by sampling other leaders, and deletions can only reduce surviving counts. Starting from empty lists, a count present in \(W\) is thus at most
\[
  \max_{V\text{ visited before or at }W}q_\star(V).
\]
Every such set contains \(W\) and, on \(\mathcal E_{\mathrm R}\), contains arm \(1\). By \eqref{app:finite:nested-gaps},
\begin{align*}
  q_\star(V) &<2B(V) \le2+\frac{288\gamma^4\Lambda_{\mathrm R}}{\eta(V)^2}\\
  &\le2+\frac{288\gamma^6\Lambda_{\mathrm R}}{\eta(W)^2} =2+288A(W) \le d(W).
\end{align*}
This controls all inherited and newly collected records by one historical maximum.
\end{proof}

\begin{lemma}[Records removed at an elimination]
\label{app:finite:deleted-records}
There is an event \(\mathcal F_{\mathrm R}\) with \(\Pr(\mathcal F_{\mathrm R})\ge1-1/(4T)\) such that, on \(\mathcal E_{\mathrm R}\cap\mathcal F_{\mathrm R}\), every elimination from \(W\) removes at most \(D(W)\) records. If the terminal active set has size at least two, at most \(|W|d(W)\) records remain.
\end{lemma}

\begin{proof}
Fix a set \(W\) containing arm \(1\), with \(m=|W|\ge2\), and fix \(j\in W\). Define
\[
  Z_{W,j}= \sum_{i\in W\setminus\{j\}} \sum_{s=1}^{d(W)}\mathbf{1}\{J_{i,W,s}=j\}.
\]
For this fixed set, all summands are independent Bernoulli variables with mean \(1/m\). Thus \(\mathbb E[Z_{W,j}]=(m-1)d(W)/m\le d(W)\), and
\begin{align*}
  \mathbb E[e^{Z_{W,j}}] &=\left(1+\frac{e-1}{m}\right)^{(m-1)d(W)}\\
  &\le\exp((e-1)d(W))\le e^{2d(W)}.
\end{align*}
Markov's inequality gives \(\Pr(Z_{W,j}>3d(W))\le e^{-d(W)}\). Let \(\mathcal F_{\mathrm R}\) be the event that \(Z_{W,j}\le3d(W)\) for every such fixed pair \((W,j)\). There are at most \(K2^{K-1}\) pairs. Since \(d(W)\ge\Lambda_{\mathrm R}\), \eqref{app:finite:uniform-prefix-probability} implies
\[
  \Pr(\mathcal F_{\mathrm R}^c) \le K2^{K-1}e^{-\Lambda_{\mathrm R}} \le\frac1{4T}.
\]
No conditioning on the realized active set is used here.

If arm \(j\) is eliminated from a realized set \(W\), its own leader list has at most \(d(W)\) records by Lemma~\ref{app:finite:counts}. For each other leader, the list is a filtered prefix of length at most \(d(W)\), by Lemmas~\ref{app:finite:prefix} and~\ref{app:finite:counts}. All its records against \(j\) are therefore counted by \(Z_{W,j}\). On \(\mathcal F_{\mathrm R}\), at most \(3d(W)\) such opponent records are deleted in total. These are disjoint from \(j\)'s leader records, giving the bound \(4d(W)=D(W)\). The terminal assertion follows by summing the per-leader bound over the terminal set.
\end{proof}

\subsubsection{Regret under relaxed transitivity}

\begin{proof}[Regret assertion of Theorem~\ref{thm:finite-relaxed}]
Work on \(\mathcal E_{\mathrm R}\cap\mathcal F_{\mathrm R}\), which has probability at least \(1-1/(2T)\). Arm \(1\) survives throughout. Each exploration comparison creates exactly one record. Charge that comparison when its first endpoint is eliminated; if neither endpoint is eliminated, charge it to the terminal set. This partitions the comparison records without duplication.

Suppose an elimination occurs from a set \(W\) of size \(m\ge2\). Both endpoints of every record deleted at this elimination belong to \(W\), even if the comparison was played earlier. By \eqref{app:finite:nested-gaps}, each endpoint has gap at most \(\gamma\eta(W)\). Thus the regret charged to this elimination is at most
\begin{align*}
  \gamma\eta(W)D(W) &\le1204\gamma\eta(W)A(W)\\
  &=\frac{1204\gamma^7\Lambda_{\mathrm R}}{\eta(W)}.
\end{align*}
Since a set of size \(m\) has worst-ranked index at least \(m\), relaxed transitivity gives \(\epsilon_m\le\gamma\eta(W)\), including equality of the two indices as an endpoint case. Hence this charge is at most \(1204\gamma^8\Lambda_{\mathrm R}/\epsilon_m\). Every elimination removes exactly one arm, so its pre-elimination size is a distinct member of \(\{2,\ldots,K\}\). The sum of all elimination charges is at most
\begin{equation}
\label{app:finite:relaxed-elimination-charge}
  1204\gamma^8\Lambda_{\mathrm R}H_{\mathrm p}.
\end{equation}

If a singleton is reached, its arm is \(1\). Any remaining records and all subsequent comparisons then have zero regret. Otherwise, let \(W_*\) be the terminal set, with \(m_*=|W_*|\ge2\) and \(\eta_* =\eta(W_*)\). The charge of its remaining records is at most
\[
  \gamma\eta_*m_*d(W_*) \le\frac{301m_*\gamma^7\Lambda_{\mathrm R}}{\eta_*}.
\]
For each \(i\in W_*\setminus\{1\}\), \(\epsilon_i\le\gamma\eta_*\). Using \(m_*\le2(m_*-1)\), we obtain
\[
  \frac{m_*}{\eta_*} \le2\gamma\sum_{i\in W_*\setminus\{1\}} \frac1{\epsilon_i}.
\]
The terminal charge is therefore at most \(602\gamma^8\Lambda_{\mathrm R}H_{\mathrm p}\). Combining with \eqref{app:finite:relaxed-elimination-charge} proves the explicit high-probability bound
\begin{equation}
\label{app:finite:relaxed-explicit-regret}
  R_T\le1806\gamma^8\Lambda_{\mathrm R}H_{\mathrm p}.
\end{equation}

Finally, \(\rho\ge1/(2T)\) implies \(\delta_{\mathrm R}\le\rho_{\mathrm R}/2\), so
\[
  \Lambda_{\mathrm R} \le\frac{128M^2\log(4MT)}{\rho^2}.
\]
On the exceptional event, \(R_T\le T/2\). Its contribution to expectation is at most \(1/4\). Thus
\[
  \mathbb E[R_T] \le1806\gamma^8\Lambda_{\mathrm R}H_{\mathrm p} +\frac14,
\]
which gives the regret bound in \eqref{eq:finite-relaxed-regret}, after absorbing universal constants and using \(M=\mathcal O(K^2\log T)\).
\end{proof}

\subsubsection{Reference execution and trajectory agreement}

Define an auxiliary execution with the same opponent arrays, rounding offsets, count-based sampling, and deletion rules as Algorithm~\ref{alg:finite-relaxed}. At stage \(\ell\), set \(W\), and level \(q\), replace every empirical rounded score by
\begin{equation}
\label{app:finite:population-reference}
  p_{\ell,q,i}^{\circ}(W) =Q_{\tau(q)/2}(p_i(W),U_{\ell,q,i}).
\end{equation}
This reference execution needs only opponent labels for its record counts and deletions.

Its potential sequence of active sets can be constructed without observing any opponent labels or outcomes. Set \(W_1^{\circ}=[K]\). Given a nonsingleton \(W_\ell^{\circ}\), inspect its population-rounded checks at every \(q\in\mathcal Q\) in increasing order. If a check first passes the elimination test, delete its minimum-score arm, using the public tie-breaking rule, to define \(W_{\ell+1}^{\circ}\). Stop at a singleton or if no level passes the test. Potential sets can be defined even when the corresponding checks would occur after time \(T\). Because actual checks in each stage occur in increasing order, the time-limited reference execution visits a prefix of this potential sequence.

Whether \(W_\ell^{\circ}\) is defined, and its value when defined, depend only on offsets from stages before \(\ell\). They are independent of the full arrays \((J_{i,s},X_{i,s}^{(1)},X_{i,s}^{(2)})_{i,s}\) constructed above and of the offsets indexed by stage \(\ell\). For each defined nonsingleton set, every \(q\in\mathcal Q\), and every \(i\in W_\ell^{\circ}\), define virtual estimates
\[
  \widehat p_{\ell,q,i}^{(r),\circ} =Q_{\tau(q)/2} (\overline X_{i,W_\ell^{\circ},q}^{(r)},U_{\ell,q,i}), \qquad r\in\{1,2\}.
\]
These use the full auxiliary arrays, whether or not the corresponding samples are collected by time \(T\).

\begin{lemma}[Simultaneous reference agreement]
\label{app:finite:reference-agreement}
With probability at least \(1-\rho\), both virtual estimates equal \(p_{\ell,q,i}^{\circ}(W_\ell^{\circ})\) for every defined nonsingleton stage, level, and active arm.
\end{lemma}

\begin{proof}
Condition on all offsets from stages before \(\ell\). This determines the existence and value of \(W_\ell^{\circ}\). If the set is defined, its filtered labels remain independent and uniform on that now-fixed set. The two observation sequences share the labels, and their comparison randomness is independent conditional on the labels. The offset \(U_{\ell,q,i}\) is still uniform and independent of these sequences. Part three of Lemma~\ref{app:finite:mean} therefore bounds the probability that either virtual estimate differs from its population-rounded value by \(\rho_{\mathrm R}\).

Every defined set at stage \(\ell\) has size \(K+1-\ell\). Union bound over its arms and levels, and then average over the preceding-stage offsets. Summing over the possible nonsingleton stages bounds total failure probability by
\[
  \rho_{\mathrm R}L\sum_{m=2}^K m \le M\rho_{\mathrm R}=\rho.
\]
The proof never conditions on agreement of earlier empirical estimates or on an outcome-dependent retained set.
\end{proof}

\begin{proof}[Replicability assertion of Theorem~\ref{thm:finite-relaxed}]
Let \(\mathcal G_{\mathrm R}\) be the event in Lemma~\ref{app:finite:reference-agreement}. We show by induction over algorithmic operations that, on this event, both real executions have the same decisions and sampling times as the reference execution.

Initially, all three executions have the same active set, stage counter, checked levels, time, empty lists, and array pointers. Suppose these quantities agree before an operation, where list agreement refers to record indices and opponent labels, not to outcome values. At a sampling operation, the counts and public tie-breaking select the same leader. The common active set and array pointer produce the same opponent, after skipping the same labels. The same record index and label are appended, and time advances by one in all executions. The recorded outcomes may differ.

At a check, the common set is \(W_\ell^{\circ}\) and the common level is \(q\). Lemma~\ref{app:finite:prefix} identifies each empirical mean with the corresponding virtual prefix mean. On \(\mathcal G_{\mathrm R}\), both rounded values equal the reference value, coordinate by coordinate. The executions therefore make the same elimination decision and choose the same arm if an elimination occurs. They delete the same record indices and update the stage counter and checked levels identically.

This induction proves agreement of all ordered pairs through time \(T\). A common singleton also yields the same subsequent self-comparisons, regardless of which arm it contains. Thus trajectory agreement on \(\mathcal G_{\mathrm R}\) does not require the accuracy event used for regret. Since \(\Pr(\mathcal G_{\mathrm R})\ge1-\rho\), Algorithm~\ref{alg:finite-relaxed} is \(\rho\)-replicable.
\end{proof}

\subsection{Strong stochastic transitivity}
\label{app:finite:sst}

This subsection uses Assumptions~\ref{asmp:strict-order}
and~\ref{asmp:sst-best}. We analyze Algorithm~\ref{alg:finite-sst}, with \(H=1+\lfloor\log_2T\rfloor\), \(N=KH\), \(\rho_{\mathrm S}=\rho/N\), \(\delta_{\mathrm S}=\delta/N\), and \(\Lambda_{\mathrm S}\) from \eqref{eq:finite-sst-parameters}. In particular, \(\Lambda_{\mathrm S}>1\). At phase \(g\), write \(m_g=|W_g|\), \(n_g=2^g\), and \(\tau_g=\sqrt{\Lambda_{\mathrm S}/n_g}\). Every completed exploration phase supplies exactly \(n_g\) fresh Bernoulli observations to each leader, including Bernoulli observations for self-comparisons. The update is
\begin{equation}
\label{app:finite:sst-update}
  W_{g+1} =\left\{i\in W_g: \max_{h\in W_g}\widehat p_{h,g}-\widehat p_{i,g} <4\tau_g\right\}.
\end{equation}
The phase index is never reset after elimination.

For explicit conditioning, let \(V_{g,s}\) be the independent shared opponent uniforms, with \(s\) indexing positions in phase \(g\), and let \(U_{g,i}\) be the independent shared rounding offsets. Map each opponent uniform to an arm using the public ordering of the current active set. Independently of these variables, take private uniforms \(Z_{g,s}^{(r)}\) for the outcomes in execution \(r\). At position \(s\), use \(\mathbf{1}\{Z_{g,s}^{(r)}<P_{i,j}\}\) for the selected pair \((i,j)\). These uniforms are independent across phases, positions, and executions, and are revealed only when used. This realizes the original fresh-feedback model. Unused phase uniforms and offsets remain independent of all histories from preceding phases.

A truncated phase has no checkpoint. A phase that completes exactly at time \(T\) may perform its checkpoint, as in Algorithm~\ref{alg:finite-sst}; its newly computed set need not start another phase.

\subsubsection{Deterministic properties of the phase schedule}
\label{app:finite:sst-schedule-section}

An exploration phase is \emph{started} if at least one of its comparisons is played. Let \(G\) be the index of the last started exploration phase. Because \(T\ge K\ge2\), phase zero is started, so \(G\) is well defined. The started exploration phases are \(0,\dots,G\). Every phase \(g<G\) is completed.

\begin{lemma}[Phase schedule and estimator budget]
\label{app:finite:sst-schedule}
On every sample path:

\begin{enumerate}
\item
The active sets are nested and nonempty. Every started exploration phase has \(m_g\ge2\).

\item
Every arm leads exactly \(n_g\) times in a completed phase and at most \(n_g\) times in a truncated phase.

\item
There are at most \(H\) started exploration phases and at most \(N\) rounded-estimator calls.

\item
For every integer \(m\ge0\), \(\sum_{g=0}^{m}n_g<2n_m\). Writing \(q_g:=\sqrt{n_g}=2^{g/2}\), we also have \(\sum_{g=0}^{m}q_g<(2+\sqrt2)q_m\).
\end{enumerate}
\end{lemma}

\begin{proof}
The update \eqref{app:finite:sst-update} only removes arms. An arm attaining the maximum rounded score has score difference zero and therefore survives, since \(\tau_g>0\). Thus every generated active set is nonempty. The algorithm switches to self-comparisons before starting an exploration phase with a singleton, which proves the first claim.

In a completed phase, the cyclic leader schedule runs for \(m_gn_g\) comparisons. Each of the \(m_g\) arms is therefore selected exactly \(n_g\) times. A truncated phase is a prefix of this schedule, so no arm leads more than \(n_g\) times.

To start phase \(G\), the algorithm must complete phases \(0,\dots,G-1\) and then play at least one additional comparison. Hence
\[
  T \ge1+\sum_{g=0}^{G-1}m_g2^g \ge1+\sum_{g=0}^{G-1}2^g =2^G.
\]
Consequently, \(G\le\lfloor\log_2T\rfloor\), and at most \(G+1\le H\) exploration phases are started. Only completed phases produce estimates, with at most \(K\) estimates per phase. The total number of estimator calls is therefore at most \(KH=N\).

Finally, the geometric identities give \(\sum_{g=0}^{m}n_g=2^{m+1}-1<2n_m\) and
\[
  \sum_{g=0}^{m}q_g =\frac{2^{(m+1)/2}-1}{\sqrt2-1} <\frac{\sqrt2}{\sqrt2-1}q_m =(2+\sqrt2)q_m.
\]
\end{proof}

\subsubsection{Consequences of the preference assumptions}
\label{app:finite:sst-structure-section}

For an active set \(W\) containing \(1\), define \(d_i(W):= p_1(W)-p_i(W)\). Let \(r_i(W):= |\{h\in W:h\le i\}|\) be the rank of \(i\) within \(W\).

\begin{proof}[Proof of Lemma~\ref{lem:finite-sst-scores}]
The statement for \(i=1\) is immediate because \(\epsilon_1=d_1(W)=0\). Fix \(i>1\) and write \(m:=|W|\). For each possible opponent \(h\in W\), let \(z_h:=\epsilon_{1,h}-\epsilon_{i,h}\). Then \(d_i(W)=m^{-1}\sum_{h\in W}z_h\).

If \(h=1\), antisymmetry gives \(z_1=-\epsilon_{i,1}=\epsilon_i\). If \(h=i\), then \(z_i=\epsilon_{1,i}=\epsilon_i\).

Suppose \(1<h<i\). Antisymmetry gives \(z_h=\epsilon_{1,h}+\epsilon_{h,i}\). The triangle inequality on \((1,h,i)\) implies \(z_h\ge\epsilon_i\). Strong stochastic transitivity on the same triple gives both \(\epsilon_{1,h}\le\epsilon_i\) and \(\epsilon_{h,i}\le\epsilon_i\). Thus \(\epsilon_i\le z_h\le2\epsilon_i\).

Suppose instead that \(i<h\). Strong stochastic transitivity on \((1,i,h)\) gives \(\epsilon_{1,h}\ge\epsilon_{i,h}\), so \(z_h\ge0\). The triangle inequality on this triple gives \(\epsilon_{1,h}\le\epsilon_i+\epsilon_{i,h}\), so \(z_h\le\epsilon_i\).

Every summand is therefore between zero and \(2\epsilon_i\). Each of the \(r_i(W)\) opponents at least as good as \(i\) contributes at least \(\epsilon_i\). Averaging proves \eqref{eq:finite-sst-scores}.

For monotonicity, the case \(i=1\) follows from \(\epsilon_1=0\) and the positivity of suboptimal gaps. For \(1<i<j\), strong stochastic transitivity on \((1,i,j)\) gives \(\epsilon_i\le\epsilon_j\). Thus the worst arm in any active set has a largest best-arm gap.
\end{proof}

\subsubsection{Uniform accuracy and elimination}
\label{app:finite:sst-elimination-section}

Let \(\mathcal E_{\mathrm S}\) be the event that
\[
  |\widehat p_{i,g}-p_i(W_g)|\le\tau_g
\]
for every completed exploration phase \(g\) and every \(i\in W_g\).

\begin{lemma}[Uniform accuracy]
\label{app:finite:sst-accuracy}
The event \(\mathcal E_{\mathrm S}\) satisfies \(\Pr(\mathcal E_{\mathrm S})\ge1-\delta\).
\end{lemma}

\begin{proof}
Consider a phase \(g\) and condition on the history immediately before its first comparison. This fixes its active set, global start time, per-arm budget, and cyclic leader schedule. It also fixes whether the phase can finish before the horizon, because its nominal length is \(m_gn_g\) and there are no decisions inside the phase that change this length.

If the phase completes, all observations entering its estimators are collected after this history. For a fixed active arm, its opponents are independent uniform draws from the fixed set \(W_g\), and its comparison outcomes use fresh independent randomness. Lemma~\ref{app:finite:mean} therefore bounds the conditional probability of an inaccurate estimate by \(\delta_{\mathrm S}\).

For each \(g\in\{0,\dots,H-1\}\) and \(i\in[K]\), let \(\mathcal F_{g,i}\) be the event that phase \(g\) completes, \(i\) is active, and its estimate is inaccurate. Interpret this event as empty if the call does not occur. Conditioning on the phase-start history and then averaging gives \(\Pr(\mathcal F_{g,i})\le\delta_{\mathrm S}\).

By Lemma~\ref{app:finite:sst-schedule}, all estimator calls are included among these \(KH=N\) possible indices. A union bound yields
\[
  \Pr(\mathcal E_{\mathrm S}^c) \le \sum_{g=0}^{H-1}\sum_{i=1}^{K} \Pr(\mathcal F_{g,i}) \le KH\delta_{\mathrm S} =\delta.
\]
\end{proof}

\begin{lemma}[Preservation of the best arm]
\label{app:finite:sst-safety}
On \(\mathcal E_{\mathrm S}\), every generated active set contains \(1\). Consequently, any singleton reached on \(\mathcal E_{\mathrm S}\) consists of \(1\).
\end{lemma}

\begin{proof}
Initially, \(1\in W_0\). Suppose \(1\in W_g\) and phase \(g\) completes. Lemma~\ref{lem:finite-sst-scores} gives \(p_h(W_g)\le p_1(W_g)\) for every \(h\in W_g\). On \(\mathcal E_{\mathrm S}\),
\begin{align*}
  \widehat p_{h,g}-\widehat p_{1,g} &\le p_h(W_g)+\tau_g -p_1(W_g)+\tau_g\\
  &\le2\tau_g.
\end{align*}
Thus \(\max_{h\in W_g}\widehat p_{h,g} -\widehat p_{1,g}\le2\tau_g<4\tau_g\), so \eqref{app:finite:sst-update} retains \(1\).

A truncated phase makes no update to the active set. Induction therefore proves the claim for all generated sets.
\end{proof}

\begin{lemma}[Elimination and survival thresholds]
\label{app:finite:sst-elimination}
On \(\mathcal E_{\mathrm S}\), at the checkpoint ending any completed phase \(g\):

\begin{enumerate}
\item
If \(d_i(W_g)\ge6\tau_g\), then \(i\) is eliminated.

\item
If \(i\) survives, then \(d_i(W_g)<6\tau_g\) and \(\epsilon_i< 6m_g\tau_g/r_i(W_g)\).

\item
If \(i\) is eliminated, then \(d_i(W_g)\ge2\tau_g\) and \(\epsilon_i\ge\tau_g\).
\end{enumerate}
\end{lemma}

\begin{proof}
By Lemma~\ref{app:finite:sst-safety}, \(1\in W_g\). On \(\mathcal E_{\mathrm S}\),
\[
  \max_{h\in W_g}\widehat p_{h,g} -\widehat p_{i,g} \ge \widehat p_{1,g}-\widehat p_{i,g} \ge d_i(W_g)-2\tau_g.
\]
If \(d_i(W_g)\ge6\tau_g\), the displayed difference is at least \(4\tau_g\), which triggers elimination.

The first statement implies by contraposition that every survivor has \(d_i(W_g)<6\tau_g\). Combining this with \eqref{eq:finite-sst-scores} proves the bound on its best-arm gap.

Finally, let \(h^\star\) attain the maximum rounded score. Since the best arm maximizes the population score, accuracy implies \(\widehat p_{h^\star,g} \le p_1(W_g)+\tau_g\) and \(\widehat p_{i,g}\ge p_i(W_g)-\tau_g\). For an eliminated arm,
\[
  4\tau_g \le \widehat p_{h^\star,g}-\widehat p_{i,g} \le d_i(W_g)+2\tau_g.
\]
Hence \(d_i(W_g)\ge2\tau_g\). The upper bound \(d_i(W_g)\le2\epsilon_i\) from Lemma~\ref{lem:finite-sst-scores} then gives \(\epsilon_i\ge\tau_g\).
\end{proof}

\subsubsection{The survivor barrier}
\label{app:finite:sst-barrier-section}

For each started exploration phase, define \(\eta_g:=\max_{i\in W_g}\epsilon_i\). On \(\mathcal E_{\mathrm S}\), this is the gap of the worst arm in \(W_g\). Moreover, \(\eta_g>0\) because \(1\in W_g\), \(m_g\ge2\), and every suboptimal arm has positive gap.

Set \(q_g:=\sqrt{n_g}\), \(u_g:=\eta_g/\tau_g=\eta_gq_g/\sqrt{\Lambda_{\mathrm S}}\), \(\Phi_g:=(m_g-1)q_g\), and \(\Psi_g:=\Phi_gu_g\). For consecutive started phases, write \(a_g:=m_g/m_{g+1}\ge1\).

\begin{lemma}[Gap of the worst survivor]
\label{app:finite:sst-survivor-gap}
On \(\mathcal E_{\mathrm S}\), for every \(g<G\), \(\eta_{g+1}<6a_g\tau_g\).
\end{lemma}

\begin{proof}
Phase \(g\) is completed because phase \(g+1\) is started. Let \(w\) be the worst arm in \(W_{g+1}\). By Lemma~\ref{lem:finite-sst-scores}, \(\epsilon_w=\eta_{g+1}\).

Every arm in \(W_{g+1}\) is at least as good as \(w\) and also belongs to \(W_g\). Therefore \(r_w(W_g)\ge m_{g+1}\). Since \(w\) survives phase \(g\), Lemmas~\ref{lem:finite-sst-scores} and~\ref{app:finite:sst-elimination} give
\[
  \frac{m_{g+1}}{m_g}\eta_{g+1} \le \frac{r_w(W_g)}{m_g}\epsilon_w \le d_w(W_g) <6\tau_g.
\]
Rearranging proves the result.
\end{proof}

\begin{lemma}[Survivor barrier]
\label{app:finite:sst-barrier}
On \(\mathcal E_{\mathrm S}\), \(u_0\le\tfrac12\) and \(\Psi_0\le\tfrac12\Phi_0\). For every started phase \(g\ge1\),
\begin{equation}
\label{app:finite:sst-survivor-barrier}
  u_g<6\sqrt2\,a_{g-1}, \qquad \Psi_g\le12\Phi_{g-1}.
\end{equation}
\end{lemma}

\begin{proof}
Since \(q_0=1\), \(\eta_0<\tfrac12\), and \(\Lambda_{\mathrm S}>1\), we have \(u_0=\eta_0/\sqrt{\Lambda_{\mathrm S}}<\tfrac12\). This also gives \(\Psi_0=\Phi_0u_0\le\Phi_0/2\).

For \(g\ge1\), Lemma~\ref{app:finite:sst-survivor-gap} gives \(\eta_g<6a_{g-1}\tau_{g-1}\). Using \(\tau_{g-1}=\sqrt{\Lambda_{\mathrm S}}/q_{g-1}\) and \(q_g=\sqrt2q_{g-1}\) yields
\[
  u_g =\frac{\eta_gq_g}{\sqrt{\Lambda_{\mathrm S}}} <6a_{g-1}\frac{q_g}{q_{g-1}} =6\sqrt2\,a_{g-1}.
\]
Consequently,
\begin{align*}
  \Psi_g &=(m_g-1)q_gu_g\\
  &< 12q_{g-1}\frac{m_{g-1}(m_g-1)}{m_g}.
\end{align*}
Nestedness gives \(m_g\le m_{g-1}\), and hence \(m_{g-1}(m_g-1)\le m_g(m_{g-1}-1)\). Substitution proves \(\Psi_g\le12q_{g-1}(m_{g-1}-1)=12\Phi_{g-1}\).
\end{proof}

\subsubsection{The nominal exploration cost}
\label{app:finite:sst-cost-section}

Define \(\Sigma_g:=\sum_{i\in W_g}\epsilon_i\), \(S:=\sum_{g=0}^{G}n_g\Sigma_g\), and \(\mathcal B:=\Lambda_{\mathrm S}\sum_{i=2}^{K}\epsilon_i^{-1}\). The quantity \(S\) charges each started phase its full nominal per-arm budget, including a truncated final phase.

We now prove a deterministic bound on \(S\) on the event \(\mathcal E_{\mathrm S}\).

\paragraph{Grouping arms by their last started phase.}
For every \(i>1\), let \(G_i:=\max\{g\in\{0,\dots,G\}:i\in W_g\}\). This is well defined because every arm belongs to \(W_0\). Let \(\mathcal D_g:=\{i:i>1,\ G_i=g\}\), \(c_g:=|\mathcal D_g|\), and \(v_g:=c_gq_g\).

On \(\mathcal E_{\mathrm S}\), the sets \(\mathcal D_0,\dots,\mathcal D_G\) partition the suboptimal arms. For \(g<G\), \(\mathcal D_g=W_g\setminus W_{g+1}\). For the last started phase, \(\mathcal D_G=W_G\setminus\{1\}\), regardless of whether its checkpoint is executed. Thus
\begin{equation}
\label{app:finite:sst-exit-counts}
  m_g-1=\sum_{h=g}^{G}c_h, \qquad c_g\le m_g-1.
\end{equation}
Every suboptimal arm appears in exactly one group, so \(\sum_gv_g>0\).

\begin{lemma}[Cost and benchmark inequalities]
\label{app:finite:sst-cost-benchmark}
On \(\mathcal E_{\mathrm S}\), with all sums over \(g=0,\dots,G\),
\begin{equation}
\label{app:finite:sst-cost-sides}
  S\le2\sqrt{\Lambda_{\mathrm S}}\sum_gv_gu_g, \qquad \mathcal B\ge \sqrt{\Lambda_{\mathrm S}}\sum_g\frac{v_g}{u_g}.
\end{equation}
\end{lemma}

\begin{proof}
Nestedness implies that a suboptimal arm \(i\) is present precisely in phases \(0,\dots,G_i\). Because \(\epsilon_1=0\), we may exchange the finite sums and use Lemma~\ref{app:finite:sst-schedule} to obtain
\begin{align*}
  S &=\sum_{i=2}^{K}\epsilon_i \sum_{g=0}^{G_i}n_g\\
  &\le2\sum_{i=2}^{K}\epsilon_i n_{G_i}\\
  &=2\sum_{g=0}^{G}n_g \sum_{i\in\mathcal D_g}\epsilon_i\\
  &\le2\sum_{g=0}^{G}n_gc_g\eta_g.
\end{align*}
Since \(n_g\eta_g=\sqrt{\Lambda_{\mathrm S}}\,q_gu_g\), the final expression equals \(2\sqrt{\Lambda_{\mathrm S}}\sum_gv_gu_g\).

For the benchmark, each arm in \(\mathcal D_g\) belongs to \(W_g\) and therefore satisfies \(\epsilon_i\le\eta_g\). Hence
\begin{align*}
  \mathcal B &=\sum_{g=0}^{G} \sum_{i\in\mathcal D_g}\frac{\Lambda_{\mathrm S}}{\epsilon_i}\\
  &\ge\sum_{g=0}^{G}c_g\frac{\Lambda_{\mathrm S}}{\eta_g}\\
  &=\sqrt{\Lambda_{\mathrm S}} \sum_{g=0}^{G}\frac{v_g}{u_g},
\end{align*}
where the last equality uses \(\Lambda_{\mathrm S}/\eta_g=\sqrt{\Lambda_{\mathrm S}}\,q_g/u_g\). All denominators are positive on \(\mathcal E_{\mathrm S}\).
\end{proof}

\begin{lemma}[Summed survivor barrier]
\label{app:finite:sst-summed-barrier}
On \(\mathcal E_{\mathrm S}\), \(\sum_{g=0}^{G}v_gu_g\le43\sum_{g=0}^{G}v_g\).
\end{lemma}

\begin{proof}
By \eqref{app:finite:sst-exit-counts}, \(v_gu_g=c_gq_gu_g\le(m_g-1)q_gu_g=\Psi_g\). The survivor barrier therefore gives
\begin{align*}
  \sum_{g=0}^{G}v_gu_g &\le\sum_{g=0}^{G}\Psi_g\\
  &\le\frac12\Phi_0 +12\sum_{g=1}^{G}\Phi_{g-1}\\
  &\le\frac{25}{2}\sum_{g=0}^{G}\Phi_g.
\end{align*}
The second sum is empty when \(G=0\), and the bound remains valid.

Next, use \eqref{app:finite:sst-exit-counts} and exchange the finite sums:
\begin{align*}
  \sum_{g=0}^{G}\Phi_g &=\sum_{g=0}^{G}q_g\sum_{h=g}^{G}c_h\\
  &=\sum_{h=0}^{G}c_h\sum_{g=0}^{h}q_g\\
  &\le(2+\sqrt2)\sum_{h=0}^{G}c_hq_h\\
  &=(2+\sqrt2)\sum_{h=0}^{G}v_h.
\end{align*}
The geometric bound is from Lemma~\ref{app:finite:sst-schedule}. Combining the two displays gives
\[
  \sum_gv_gu_g \le\frac{25}{2}(2+\sqrt2)\sum_gv_g \le43\sum_gv_g,
\]
since \(\tfrac{25}{2}(2+\sqrt2)<43\).
\end{proof}

\begin{lemma}[Exploration cost]
\label{app:finite:sst-exploration-cost}
On \(\mathcal E_{\mathrm S}\), \(S\le3698\mathcal B\).
\end{lemma}

\begin{proof}
Because \(u_g>0\) and \(v_g\ge0\), Cauchy--Schwarz gives
\begin{align*}
  \left(\sum_gv_g\right)^2 &= \left( \sum_g\sqrt{v_gu_g}\sqrt{\frac{v_g}{u_g}} \right)^2\\
  &\le \left(\sum_gv_gu_g\right) \left(\sum_g\frac{v_g}{u_g}\right)\\
  &\le 43\left(\sum_gv_g\right) \left(\sum_g\frac{v_g}{u_g}\right),
\end{align*}
where the last step uses Lemma~\ref{app:finite:sst-summed-barrier}. Since \(\sum_gv_g>0\), division gives \(\sum_gv_g\le43\sum_gv_g/u_g\). Applying Lemma~\ref{app:finite:sst-summed-barrier} again, \(\sum_gv_gu_g\le43^2\sum_gv_g/u_g\).

The two inequalities in \eqref{app:finite:sst-cost-sides} now imply
\[
  S \le2\sqrt{\Lambda_{\mathrm S}}\sum_gv_gu_g \le2\cdot43^2\sqrt{\Lambda_{\mathrm S}}\sum_g\frac{v_g}{u_g} \le3698\mathcal B.
\]
\end{proof}

\subsubsection{High-probability regret}
\label{app:finite:sst-regret-section}

Let \(\mathcal I_{\mathrm{exp}}\subseteq[T]\) be the set of exploration rounds. Define the cumulative leader and opponent gaps over these rounds by \(A:=\sum_{t\in\mathcal I_{\mathrm{exp}}}\epsilon_{i_t}\) and \(D:=\sum_{t\in\mathcal I_{\mathrm{exp}}}\epsilon_{j_t}\). These quantities are defined on every sample path, including paths outside \(\mathcal E_{\mathrm S}\).

\begin{lemma}[Leader contribution]
\label{app:finite:sst-leader}
Deterministically, \(A\le S\).
\end{lemma}

\begin{proof}
In a completed phase \(g\), each arm leads exactly \(n_g\) times, giving leader contribution \(n_g\Sigma_g\). In a truncated phase, each arm leads at most \(n_g\) times. Since all gaps are nonnegative, its leader contribution is at most \(n_g\Sigma_g\). Summing over all started phases proves \(A\le S\).
\end{proof}

\begin{lemma}[Opponent contribution]
\label{app:finite:sst-opponent}
There is an event \(\mathcal H_{\mathrm S}\) with \(\Pr(\mathcal H_{\mathrm S})\ge1-\delta\) on which \(D\le2S+\log(1/\delta)\).
\end{lemma}

\begin{proof}
For every \(t\in[T]\), define \(\xi_t:=\epsilon_{j_t} \mathbf{1}\{t\in\mathcal I_{\mathrm{exp}}\}\). Thus \(D=\sum_{t=1}^{T}\xi_t\) and \(0\le\xi_t\le\tfrac12\).

Let \(\mathcal K_t\) be the information available immediately before drawing the opponent at round \(t\). It includes the past observations, all previously revealed algorithmic randomness, the current mode of the algorithm, the active set, and the leader. At an exploitation round, take \(\mathcal K_t\) to be the information immediately before that round's self-comparison. In particular, exploration status and the current active set are \(\mathcal K_t\)-measurable.

Write \(\nu_t:=\mathbb E[\xi_t\mid\mathcal K_t]\). At an exploration round in phase \(g\), the fresh opponent draw is uniform over \(W_g\), so \(\nu_t=\Sigma_g/m_g\). At an exploitation round, \(\nu_t=0\). Every phase contributes at most \(m_gn_g\) rounds, and hence, on every sample path,
\begin{equation}
\label{app:finite:sst-opponent-predictable}
  \sum_{t=1}^{T}\nu_t \le \sum_{g=0}^{G}m_gn_g\frac{\Sigma_g}{m_g} =S.
\end{equation}

For \(x\in[0,1]\), convexity gives \(e^x\le1+(e-1)x\le1+2x\). Applying this to \(\xi_t\) yields
\[
  \mathbb E[e^{\xi_t}\mid\mathcal K_t] \le1+2\nu_t \le e^{2\nu_t}.
\]
Define \(Z_t:=\exp(\sum_{s=1}^{t}(\xi_s-2\nu_s))\) and \(Z_0:=1\). All terms from rounds before \(t\) are \(\mathcal K_t\)-measurable, so
\[
  \mathbb E[Z_t\mid\mathcal K_t] = Z_{t-1}e^{-2\nu_t} \mathbb E[e^{\xi_t}\mid\mathcal K_t] \le Z_{t-1}.
\]
Iterating conditional expectations gives \(\mathbb E[Z_T]\le1\). Markov's inequality therefore implies
\[
  \Pr\!\left( D-2\sum_{t=1}^{T}\nu_t \ge\log(1/\delta) \right) = \Pr(Z_T\ge1/\delta) \le\delta.
\]
Together with \eqref{app:finite:sst-opponent-predictable}, this proves the claim.
\end{proof}

\begin{proof}[Proof of the regret assertion in Theorem~\ref{thm:finite-sst}]
By Lemmas~\ref{app:finite:sst-accuracy} and~\ref{app:finite:sst-opponent}, \(\Pr(\mathcal E_{\mathrm S}\cap\mathcal H_{\mathrm S})\ge1-2\delta\). Work on this intersection.

Lemma~\ref{app:finite:sst-safety} ensures that the best arm is never eliminated. Any exploitation rounds therefore compare \((1,1)\) and have zero regret. Consequently,
\begin{align*}
  R_T &=\frac12(A+D)\\
  &\le\frac32S+\frac12\log(1/\delta)\\
  &\le5547\mathcal B+\frac12\log(1/\delta),
\end{align*}
using Lemmas~\ref{app:finite:sst-leader}, \ref{app:finite:sst-opponent}, and \ref{app:finite:sst-exploration-cost}.

It remains to absorb the logarithmic term. Because \(\epsilon_i<\tfrac12\) for \(i>1\), \(\mathcal B\ge2\Lambda_{\mathrm S}(K-1)\ge2\Lambda_{\mathrm S}\). Also, \(0<\rho_{\mathrm S}-\delta_{\mathrm S}<1\) and \(\delta_{\mathrm S}=\delta/N\) imply \(\Lambda_{\mathrm S}\ge8\log(1/\delta)\). Thus \(\mathcal B\ge16\log(1/\delta)\), and
\[
  R_T \le \left(5547+\frac1{32}\right)\mathcal B \le6000\mathcal B = 6000\Lambda_{\mathrm S}\sum_{i=2}^{K}\frac1{\epsilon_i}.
\]
This proves the high-probability assertion together with preservation of the best arm on the stated event.
\end{proof}

\paragraph{Expected regret.}
Choose \(\delta=\min\{\rho/4,1/T\}\). Then \(0<\delta<\rho/3\), \(\delta T\le1\), and \(\rho-\delta\ge3\rho/4\). The high-probability bound and \(R_T\le T/2\) imply
\begin{align*}
  \mathbb E[R_T] &\le6000\Lambda_{\mathrm S}H_{\mathrm p}+\delta T\\
  &\le6000\Lambda_{\mathrm S}H_{\mathrm p}+1.
\end{align*}
Moreover,
\begin{align*}
  \Lambda_{\mathrm S} &=\frac{8N^2\log(4N/\delta)}{(\rho-\delta)^2}\\
  &\le\frac{128N^2}{9\rho^2} \log\!\left(4N\max\{4/\rho,T\}\right).
\end{align*}
Since \(N=K(1+\lfloor\log_2T\rfloor)\), this proves \eqref{eq:finite-sst-regret}.

\subsubsection{Trajectory replicability}
\label{app:finite:sst-coupling-section}

Consider two executions using the shared opponent uniforms \(V_{g,s}\) and rounding offsets \(U_{g,i}\) specified in Section~\ref{app:finite:sst}, with independent comparison randomness. We bound the probability of the first disagreement between their rounded score vectors.

For this analysis, track the executions jointly until their rounded vectors first disagree. At a common phase start, let \(\mathcal H_g\) denote the \(\sigma\)-field generated by both histories and all shared randomness revealed before phase \(g\). If tracking has already stopped, use the stopped history for \(\mathcal H_g\). Unused opponent uniforms and rounding offsets are excluded from this \(\sigma\)-field.

Let \(\mathcal M_g\) be the event that both executions reach phase \(g\) with the same active set \(W_g\), the same global time, and agreement of all earlier rounded vectors, and that phase \(g\) will complete. This event is \(\mathcal H_g\)-measurable: if \(t_g\) comparisons have already been played, completion is determined by \(t_g+|W_g|n_g\le T\).

On \(\mathcal M_g\), write \(\widehat{\boldsymbol p}_g^{(r)} :=(\widehat p_{i,g}^{(r)})_{i\in W_g}\) for execution \(r\)'s rounded score vector, with coordinates arranged in the fixed public arm order. Define the first-disagreement event
\[
  \mathcal L_g := \mathcal M_g \cap \left\{ \widehat{\boldsymbol p}_g^{(1)} \ne \widehat{\boldsymbol p}_g^{(2)} \right\}.
\]
The event is empty when a common completed phase \(g\) is not reached.

\begin{lemma}[First checkpoint disagreement]
\label{app:finite:sst-phase-coupling}
For every \(g\in\{0,\dots,H-1\}\),
\begin{equation}
\label{app:finite:sst-checkpoint-disagreement}
  \Pr(\mathcal L_g\mid\mathcal H_g) \le K\rho_{\mathrm S}\,\mathbf{1}_{\mathcal M_g} \quad\text{almost surely}.
\end{equation}
Consequently, \(\Pr(\mathcal L_g)\le K\rho_{\mathrm S}\).
\end{lemma}

\begin{proof}
On \(\mathcal M_g^c\), the event \(\mathcal L_g\) is empty and its conditional probability is zero. It remains to establish the bound on \(\mathcal M_g\).

Fix the two pre-phase histories on \(\mathcal M_g\). They determine the same active set, phase index, global start time, and cyclic leader schedule. Throughout the phase, pair selection uses this schedule and the fresh opponent uniforms, without using the phase's observed outcomes. The shared uniforms therefore produce the same ordered opponent labels in both executions.

Let \(\boldsymbol J_g\) denote the collection of opponent labels in this phase, and condition additionally on \(\boldsymbol J_g\). Each active arm then has exactly \(n_g\) leading observations against the same ordered opponent sequence in both executions.

These observations use fresh comparison randomness. Conditional on \(\mathcal H_g\) and \(\boldsymbol J_g\), they are independent across comparisons and executions, with the Bernoulli means specified by the corresponding opponent labels. The reset ensures that no observation from the conditioned pre-phase histories enters a current empirical mean.

For each \(i\in W_g\), the shared offset \(U_{g,i}\) is independent of \(\mathcal H_g\), \(\boldsymbol J_g\), and the current comparison outcomes. The shared-opponent guarantee in Lemma~\ref{app:finite:mean} thus gives, on \(\mathcal M_g\),
\[
  \Pr\!\left( \widehat p_{i,g}^{(1)} \ne\widehat p_{i,g}^{(2)} \,\middle|\, \mathcal H_g,\boldsymbol J_g \right) \le\rho_{\mathrm S}.
\]
A union bound over the \(m_g=|W_g|\) active arms yields
\[
  \Pr\!\left( \widehat{\boldsymbol p}_g^{(1)} \ne\widehat{\boldsymbol p}_g^{(2)} \,\middle|\, \mathcal H_g,\boldsymbol J_g \right) \le m_g\rho_{\mathrm S} \le K\rho_{\mathrm S} \qquad\text{on }\mathcal M_g.
\]
Averaging over the fresh opponent labels proves \eqref{app:finite:sst-checkpoint-disagreement}. Taking expectations then gives \(\Pr(\mathcal L_g)\le K\rho_{\mathrm S}\).
\end{proof}

\begin{proof}[Proof of the replicability assertion in Theorem~\ref{thm:finite-sst}]
We first show that agreement of the rounded vectors at every completed checkpoint implies agreement of the entire action trajectories.

Initially, both executions have active set \([K]\), phase index zero, and global time zero. Suppose they have reached the same phase index with the same active set and global time.

If the active set is a singleton, both executions compare its unique arm with itself for all remaining rounds. Otherwise, the fixed cyclic order gives the same leader at every position of the phase. The same opponent uniform, applied to the same ordered active set, gives the same opponent. Hence every ordered pair played during the phase agrees, regardless of the observed comparison outcomes.

The common active-set size and phase index give the same nominal phase length. Since the start times also agree, either both executions complete the phase or the horizon truncates both at the same position. In the latter case, their pairs agree through the horizon and no further checkpoint is executed.

If the phase completes and its rounded vectors agree, the elimination rule \eqref{app:finite:sst-update} produces the same next active set. Both executions have played the same number of comparisons and advance to the same next phase index. This proves the induction.

Consequently, a disagreement between the action trajectories requires a first disagreement between rounded vectors at a common completed checkpoint. At that checkpoint, all earlier rounded vectors agree, so the corresponding event \(\mathcal L_g\) occurs. Therefore
\[
  \left\{ \exists t\in[T]: (i_t^{(1)},j_t^{(1)}) \ne(i_t^{(2)},j_t^{(2)}) \right\} \subseteq \bigcup_{g=0}^{H-1}\mathcal L_g.
\]
Lemma~\ref{app:finite:sst-schedule} bounds the possible phase indices by \(0,\dots,H-1\). Applying Lemma~\ref{app:finite:sst-phase-coupling} and a union bound gives
\begin{align*}
  &\Pr\!\left( \exists t\in[T]: (i_t^{(1)},j_t^{(1)}) \ne(i_t^{(2)},j_t^{(2)}) \right)\\
  &\qquad\le \sum_{g=0}^{H-1}\Pr(\mathcal L_g) \le HK\rho_{\mathrm S} =N\rho_{\mathrm S} =\rho.
\end{align*}
Thus the two executions select identical ordered pairs at every round with probability at least \(1-\rho\), as required.
\end{proof}

%% file: appendix/linear.tex
This appendix specifies the algorithm and proves Theorem~\ref{thm:repmaxinp}. The indexed feature sequence is fixed before the interaction and is the same in both executions. All probabilities include the shared random shifts and the comparison-noise variables.

\subsection{Model and algorithm specification}
\label{app:repmaxinp-specification}

For the selected pair, write \(x_t=x_{t,i_t}\), \(y_t=x_{t,j_t}\), and \(z_t=x_t-y_t\). We make explicit the model conditions from Section~\ref{sec:repmaxinp}.

\begin{assumption}[Contexts and parameter space]
\label{ass:repmaxinp-contexts}
The indexed feature sequence is fixed before the interaction and revealed online, with \(K_t\ge1\). There are known \(L,S>0\) and a known compact convex set \(\Theta\subset\mathbb R^d\) containing \(0\) such that \(\|x_{t,i}\|_2\le L\), \(\sup_{\theta\in\Theta}\|\theta\|_2\le S\), and \(\theta^\star\in\Theta\).
\end{assumption}

\begin{assumption}[Preference observations]
\label{ass:repmaxinp-feedback}
The known link \(\mu:\mathbb R\to(0,1)\) is continuously differentiable, increasing, and satisfies \(\mu(-u)=1-\mu(u)\). The binary feedback obeys \(\mathbb E[O_t\mid\mathcal F_t^{-}]=\mu(z_t^\top\theta^\star)\). For known \(R>0\), the noise \(\varepsilon_t:=O_t-\mu(z_t^\top\theta^\star)\) satisfies \(\mathbb E[e^{a\varepsilon_t}\mid\mathcal F_t^{-}]\le e^{R^2a^2/2}\) for every \(a\in\mathbb R\); \(R=1/2\) is valid for binary feedback.
\end{assumption}

\begin{assumption}[Link curvature]
\label{ass:repmaxinp-link}
There is a known \(c_\mu\in(0,1]\) such that \(\mu'(z^\top\theta)\ge c_\mu\) whenever \(\theta\in\Theta\) and \(\|z\|_2\le2L\).
\end{assumption}

Utility regret is defined by \eqref{eq:utility-regret}; in particular, \(0\le r_t^{\mathrm{util}}\le2LS\). We retain the notation \(R_T^{\mathrm{util}}\) to distinguish it from probability regret. The binary feedback permits \(R=1/2\), which is the choice used in the main theorem. The formulas below also hold for any supplied valid sub-Gaussian parameter \(R>0\).

\paragraph{Replicable parameter estimates.}
Fix \(\lambda>0\), \(\rho\in(0,1)\), and \(0<\delta<\rho/3\). Define
\begin{equation}\label{eq:repmaxinp-batch-budget}
D_{\max}:=d\log\left(1+\frac{4TL^2}{\lambda d}\right), \qquad B:=\max\{1,\lceil D_{\max}\rceil\}.
\end{equation}
Allocate \(\eta:=\delta/B\) and \(\varrho:=\rho/B\) to each estimator call. Let \(V_t:=\lambda I_d+\sum_{s<t}z_sz_s^\top\), write \(\|v\|_V:=\sqrt{v^\top Vv}\), and choose \(\psi\) with \(\psi'=\mu\). The regularized estimate and its statistical radius are
\begin{align}
\widehat\theta_t&:=\operatorname*{arg\,max}_{\theta\in\Theta}\left\{\sum_{s<t}\bigl[O_sz_s^\top\theta-\psi(z_s^\top\theta)\bigr]-\frac{\lambda}{2}\|\theta\|_2^2\right\}, \label{eq:repmaxinp-estimator}\\
\beta_t(\eta)&:=\frac{1}{c_\mu}\left[R\sqrt{\log\frac{\det(V_t)}{\det(\lambda I_d)}+2\log\frac1\eta}+\sqrt{\lambda}S\right]. \label{eq:repmaxinp-beta}
\end{align}
At batch start \(\tau_b\), reveal a fresh shared \(W_b\sim\operatorname{Unif}([0,1)^d)\), independent of earlier shifts and comparison noise. The \textnormal{RepGLM} estimate is defined by
\begin{align}
\alpha_b&:=\frac{2\beta_{\tau_b}(\eta)\sqrt d}{\varrho-2\eta}, \qquad \Gamma_b:=\beta_{\tau_b}(\eta)\left(1+\frac{d}{\varrho-2\eta}\right), \label{eq:repmaxinp-radii}\\
[\mathsf Q_{\alpha,w}(v)]_\ell&:=\alpha\left(\left\lfloor\frac{v_\ell}{\alpha}-w_\ell\right\rfloor+w_\ell+\frac12\right), \qquad \ell\in[d], \label{eq:repmaxinp-quantizer}\\
\widetilde\theta_b&:=V_{\tau_b}^{-1/2}\mathsf Q_{\alpha_b,W_b}\bigl(V_{\tau_b}^{1/2}\widehat\theta_{\tau_b}\bigr). \label{eq:repmaxinp-rounding}
\end{align}
Matrix square roots are symmetric and positive definite. Rounding in the transformed coordinates controls error in the \(V_{\tau_b}\) norm; \(\Gamma_b\) includes this rounding error. The rounded vector need not belong to \(\Theta\): the curvature assumption is used only along the segment joining the constrained estimate \(\widehat\theta_{\tau_b}\) to \(\theta^\star\), whereas the rounding analysis controls the additional error directly.

During batch \(b\), the promising set is defined by \eqref{eq:repmaxinp-promising}. This set contains an estimated-utility maximizer and is therefore nonempty. Algorithm~\ref{alg:repmaxinp} selects its widest pair and starts a new batch after sufficient determinant growth. All maximizations over indexed actions use lexicographic tie-breaking, including when distinct indices have identical features.

\begin{algorithm}[t]
\caption{$\RepMaxInP$: replicable linear dueling bandits}
\label{alg:repmaxinp}
\begin{algorithmic}[1]
\REQUIRE \(T\ge1\), \(\lambda>0\), \(\rho\in(0,1)\), \(0<\delta<\rho/3\), and known model quantities.
\STATE Compute \(D_{\max},B\) from \eqref{eq:repmaxinp-batch-budget}; \(\eta\gets\delta/B\); \(\varrho\gets\rho/B\).
\STATE \(V_1\gets\lambda I_d\); \(b\gets0\); \(\tau_0\gets1\); initialize empty records.
\FOR{\(t=1,\ldots,T\)}
  \STATE Observe the indexed features \(x_{t,1},\ldots,x_{t,K_t}\).
  \IF{\(t=\tau_b\)}
    \STATE Fit \(\widehat\theta_t\) using all past data; reveal fresh shared \(W_b\sim\operatorname{Unif}([0,1)^d)\).
    \STATE Compute \(\alpha_b,\Gamma_b,\widetilde\theta_b\) using \eqref{eq:repmaxinp-radii}--\eqref{eq:repmaxinp-rounding}.
  \ENDIF
  \STATE Form \(\mathcal C_t\) using \eqref{eq:repmaxinp-promising}.
  \STATE \((i_t,j_t)\gets\operatorname*{arg\,max}_{(i,j)\in\mathcal C_t^2}\|x_{t,i}-x_{t,j}\|_{V_{\tau_b}^{-1}}\), with lexicographic tie-breaking.
  \STATE Play \((i_t,j_t)\) and observe \(O_t\); \(z_t\gets x_{t,i_t}-x_{t,j_t}\).
  \STATE Retain \((z_t,O_t)\); \(V_{t+1}\gets V_t+z_tz_t^\top\).
  \IF{\(t<T\) and \(\det(V_{t+1})>e\det(V_{\tau_b})\)}
    \STATE \(b\gets b+1\); \(\tau_b\gets t+1\).
  \ENDIF
\ENDFOR
\end{algorithmic}
\end{algorithm}

Define the uniform radius bound
\begin{equation}\label{eq:repmaxinp-gamma-bar}
\overline\Gamma:=\frac{1}{c_\mu}\left[R\sqrt{D_{\max}+2\log(B/\delta)}+\sqrt{\lambda}S\right]\left(1+\frac{dB}{\rho-2\delta}\right).
\end{equation}

\subsection{Explicit regret guarantee}
\label{app:repmaxinp-explicit}

The following statement gives the finite-time bound underlying Theorem~\ref{thm:repmaxinp}.

\begin{theorem}[Explicit guarantee for $\RepMaxInP$]
\label{thm:repmaxinp-explicit}
Under Assumptions~\ref{ass:repmaxinp-contexts}--\ref{ass:repmaxinp-link}, for every \(T\ge1\), \(\lambda>0\), \(\rho\in(0,1)\), and \(0<\delta<\rho/3\), Algorithm~\ref{alg:repmaxinp} uses at most \(B\) batches and is \(\rho\)-replicable in the sense of Definition~\ref{def:replicability}. With probability at least \(1-\delta\), every utility-maximizing index belongs to \(\mathcal C_t\) at every round and
\begin{equation}
\label{eq:repmaxinp-hp}
R_T^{\mathrm{util}}\le2\max\{LS,\overline\Gamma\}\sqrt{2eTD_{\max}}.
\end{equation}
Moreover,
\begin{equation}
\label{eq:repmaxinp-expectation}
\mathbb E[R_T^{\mathrm{util}}]\le2\max\{LS,\overline\Gamma\}\sqrt{2eTD_{\max}}+2LST\delta.
\end{equation}
For fixed \(L,S,R,\lambda\), the high-probability rate is \(\widetilde{\mathcal O}(d^3\sqrt T/(c_\mu\rho))\). The same rate holds in expectation for \(\delta=\min\{\rho/4,1/T\}\).
\end{theorem}

\subsection{Notation and probability model}
\label{app:repmaxinp-setting}

Recall that \(z_t=x_{t,i_t}-x_{t,j_t}\), \(V_t=\lambda I_d+\sum_{s<t}z_sz_s^\top\), and \(\varepsilon_t=O_t-\mu(z_t^\top\theta^\star)\). The feature bound gives \(\|z_t\|_2\le2L\). Let \(\mathcal F_t^{-}\) be the pre-feedback sigma-field from the preliminaries, and let \(\mathcal F_t:=\sigma(\mathcal F_t^{-},O_t)\) be the post-feedback sigma-field. These satisfy \(\mathcal F_{t-1}\subseteq\mathcal F_t^{-}\subseteq\mathcal F_t\). In particular, \(z_t\) is \(\mathcal F_t^{-}\)-measurable, while \(V_t\) depends only on comparisons selected before round \(t\).

The deterministic budgets are \(B=\max\{1,\lceil D_{\max}\rceil\}\), \(\eta=\delta/B\), and \(\varrho=\rho/B\), where \(D_{\max}=d\log(1+4TL^2/(\lambda d))\). Since \(0<\delta<\rho/3\), we have \(\varrho-2\eta=(\rho-2\delta)/B>0\). Thus every grid width and confidence radius in \eqref{eq:repmaxinp-radii} is well defined.

At the start \(\tau_b\) of batch \(b\), the algorithm computes its parameter estimate from all observations before \(\tau_b\). It then reveals the independent shared shift \(W_b\) and freezes \(\widetilde\theta_b\), \(\Gamma_b\), and \(V_{\tau_b}\) for pair selection throughout that batch. The matrix \(V_t\) continues to accumulate the selected comparison features and determines when the next batch begins.

For trajectory replicability, we use two independent comparison-noise streams and a common sequence of shifts. A concrete realization is to take independent variables \(Z_t^{(k)}\sim\operatorname{Unif}[0,1)\) for \(t\in[T]\) and \(k\in\{1,2\}\), independent of all shifts, and generate the feedback in execution \(k\) as
\begin{equation}\label{app:eq:repmaxinp-feedback-realization}
O_t^{(k)}:=\mathbf 1\left\{Z_t^{(k)}<\mu\bigl((z_t^{(k)})^\top\theta^\star\bigr)\right\}.
\end{equation}
This realizes the stipulated conditional Bernoulli law. The two executions share \(W_0,\ldots,W_{B-1}\) and use the same deterministic rules for optimization and index-based tie-breaking.

\subsection{Anytime confidence and randomized rounding}
\label{app:repmaxinp-estimation}

We first establish a simultaneous confidence bound for the unrounded estimator. We then prove the grid-rounding guarantee and combine the two results.

\begin{lemma}[Self-normalized concentration]
\label{app:lem:repmaxinp-self-normalized}
Define \(S_t:=\sum_{s<t}\varepsilon_s z_s\) for \(t\le T+1\). For every \(\eta\in(0,1)\), with probability at least \(1-\eta\),
\begin{equation}\label{app:eq:repmaxinp-self-normalized}
\|S_t\|_{V_t^{-1}}\le R\sqrt{\log\frac{\det(V_t)}{\det(\lambda I_d)}+2\log\frac1\eta}
\qquad\text{for every }t\le T+1.
\end{equation}
\end{lemma}

\begin{proof}
Fix \(u\in\mathbb R^d\) and define
\begin{equation}\label{app:eq:repmaxinp-exponential-process}
M_t(u):=\exp\left(u^\top S_t-\frac{R^2}{2}u^\top(V_t-\lambda I_d)u\right).
\end{equation}
Since \(S_{t+1}=S_t+\varepsilon_tz_t\) and \(V_{t+1}=V_t+z_tz_t^\top\), conditional sub-Gaussianity gives
\begin{align*}
\mathbb E[M_{t+1}(u)\mid\mathcal F_t^{-}]
&=M_t(u)\,\mathbb E\left[\exp\left(\varepsilon_tu^\top z_t-\frac{R^2}{2}(u^\top z_t)^2\right)\middle|\mathcal F_t^{-}\right]\\
&\le M_t(u).
\end{align*}
Taking another conditional expectation given \(\mathcal F_{t-1}\) shows that \((M_t(u))_{t=1}^{T+1}\) is a nonnegative supermartingale with respect to the shifted filtration \((\mathcal F_{t-1})_{t=1}^{T+1}\). Its initial value is \(M_1(u)=1\).

Integrate these supermartingales against the Gaussian distribution \(\nu=\mathcal N(0,(R^2\lambda)^{-1}I_d)\), and write \(\overline M_t:=\int M_t(u)\,\nu(du)\). Nonnegativity permits exchanging integration and conditional expectation, so \((\overline M_t)\) is also a nonnegative supermartingale with \(\overline M_1=1\). Completing the square gives
\begin{align*}
\overline M_t
&=\left(\frac{R^2\lambda}{2\pi}\right)^{d/2}\int_{\mathbb R^d}\exp\left(u^\top S_t-\frac{R^2}{2}u^\top V_tu\right)\,du\\
&=\frac{\det(\lambda I_d)^{1/2}}{\det(V_t)^{1/2}}\exp\left(\frac{\|S_t\|_{V_t^{-1}}^2}{2R^2}\right).
\end{align*}
For completeness, the square is centered at \(R^{-2}V_t^{-1}S_t\), and its constant term is \(\|S_t\|_{V_t^{-1}}^2/(2R^2)\). Integrating the remaining Gaussian contributes \((2\pi)^{d/2}\det(R^2V_t)^{-1/2}\), yielding the displayed determinant ratio.

Let \(\sigma\) be the first time \(t\le T+1\) at which \(\overline M_t\ge1/\eta\), with \(\sigma=T+1\) if no such time exists. The stopped process is a nonnegative supermartingale, so \(\mathbb E[\overline M_\sigma]\le1\). On the event that the threshold is reached, \(\overline M_\sigma\ge1/\eta\). Consequently,
\begin{equation}\label{app:eq:repmaxinp-mixture-crossing}
\Pr\left(\max_{t\le T+1}\overline M_t\ge\frac1\eta\right)\le\eta.
\end{equation}
Outside this event, taking logarithms of the expression for \(\overline M_t\) gives
\[
\frac{\|S_t\|_{V_t^{-1}}^2}{2R^2}<\frac12\log\frac{\det(V_t)}{\det(\lambda I_d)}+\log\frac1\eta
\qquad\text{for every }t\le T+1.
\]
Rearranging proves \eqref{app:eq:repmaxinp-self-normalized}.
\end{proof}

\begin{lemma}[Deterministic estimation bound]
\label{app:lem:repmaxinp-estimation-bound}
For every realized data sequence and every \(t\le T+1\), the estimator in \eqref{eq:repmaxinp-estimator} satisfies
\begin{equation}\label{app:eq:repmaxinp-estimation-bound}
\|\widehat\theta_t-\theta^\star\|_{V_t}\le\frac{1}{c_\mu}\left(\|S_t\|_{V_t^{-1}}+\sqrt{\lambda}S\right).
\end{equation}
\end{lemma}

\begin{proof}
Write the maximized objective as \(\ell_t(\theta):=\sum_{s<t}[O_sz_s^\top\theta-\psi(z_s^\top\theta)]-\lambda\|\theta\|_2^2/2\). Since \(\psi'=\mu\) and \(\mu\) is increasing, \(\psi\) is convex. The term \(-\lambda\|\theta\|_2^2/2\) makes \(\ell_t\) strictly concave. Continuity and compactness of \(\Theta\) give existence of a maximizer, and strict concavity on the convex set \(\Theta\) gives uniqueness.

Define \(g_t(\theta):=\lambda\theta+\sum_{s<t}\mu(z_s^\top\theta)z_s\) and \(\Delta_t:=\widehat\theta_t-\theta^\star\). The segment \(\theta^\star+u\Delta_t\), \(u\in[0,1]\), lies in \(\Theta\). The fundamental theorem of calculus therefore gives
\begin{align*}
\Delta_t^\top\bigl(g_t(\widehat\theta_t)-g_t(\theta^\star)\bigr)
&=\int_0^1\Delta_t^\top\left[\lambda I_d+\sum_{s<t}\mu'\bigl(z_s^\top(\theta^\star+u\Delta_t)\bigr)z_sz_s^\top\right]\Delta_t\,du\\
&\ge c_\mu\,\|\Delta_t\|_{V_t}^2.
\end{align*}
The inequality uses \(\|z_s\|_2\le2L\), the link-curvature assumption along the segment, and \(c_\mu\le1\) for the regularization term.

The first-order optimality condition for the constrained maximizer is \(\nabla\ell_t(\widehat\theta_t)^\top(\theta-\widehat\theta_t)\le0\) for every \(\theta\in\Theta\). Taking \(\theta=\theta^\star\) gives \(\Delta_t^\top\nabla\ell_t(\widehat\theta_t)\ge0\). Also,
\[
\nabla\ell_t(\widehat\theta_t)=S_t-\lambda\theta^\star-\bigl(g_t(\widehat\theta_t)-g_t(\theta^\star)\bigr).
\]
Combining these relations yields
\begin{align*}
c_\mu\,\|\Delta_t\|_{V_t}^2
&\le\Delta_t^\top\bigl(g_t(\widehat\theta_t)-g_t(\theta^\star)\bigr)\\
&\le\Delta_t^\top(S_t-\lambda\theta^\star)\\
&\le\|\Delta_t\|_{V_t}\left(\|S_t\|_{V_t^{-1}}+\lambda\|\theta^\star\|_{V_t^{-1}}\right).
\end{align*}
Because \(V_t\succeq\lambda I_d\), we have \(\lambda\|\theta^\star\|_{V_t^{-1}}\le\sqrt{\lambda}\|\theta^\star\|_2\le\sqrt{\lambda}S\). If \(\Delta_t=0\), the claimed bound is immediate. Otherwise, dividing by \(c_\mu\|\Delta_t\|_{V_t}\) proves \eqref{app:eq:repmaxinp-estimation-bound}.
\end{proof}

\begin{lemma}[Shared-grid rounding]
\label{lem:repmaxinp-grid}
For fixed \(v,v'\in\mathbb R^d\) and \(\alpha>0\), \(\|\mathsf Q_{\alpha,w}(v)-v\|_2\le\alpha\sqrt d/2\) for every \(w\in[0,1)^d\). A shared uniform \(W\) satisfies \(\Pr(\mathsf Q_{\alpha,W}(v)\ne \mathsf Q_{\alpha,W}(v'))\le\|v-v'\|_1/\alpha\le\sqrt d\,\|v-v'\|_2/\alpha\).
\end{lemma}

\begin{proof}[Proof of Lemma~\ref{lem:repmaxinp-grid}]
Fix a coordinate \(\ell\) and let \(k=\lfloor v_\ell/\alpha-w_\ell\rfloor\). Then \(v_\ell\) lies in the interval \([\alpha(k+w_\ell),\alpha(k+w_\ell+1))\), whose midpoint is \([\mathsf Q_{\alpha,w}(v)]_\ell\). Thus \(|[\mathsf Q_{\alpha,w}(v)]_\ell-v_\ell|\le\alpha/2\). Summing the squared errors over all coordinates gives \(\|\mathsf Q_{\alpha,w}(v)-v\|_2\le\alpha\sqrt d/2\).

For the disagreement bound, first consider scalar values \(a\le b\) and a scalar shift \(U\sim\operatorname{Unif}[0,1)\). The rounded values can differ only if a grid boundary \(\alpha(k+U)\) lies in \((a,b]\) for some \(k\in\mathbb Z\). Define the number of such boundaries as \(N(U):=\sum_{k\in\mathbb Z}\mathbf 1\{a<\alpha(k+U)\le b\}\). By nonnegativity and a change of variables,
\begin{align*}
\mathbb E[N(U)]
&=\int_0^1\sum_{k\in\mathbb Z}\mathbf 1\{a/\alpha<k+u\le b/\alpha\}\,du\\
&=\int_{\mathbb R}\mathbf 1\{a/\alpha<v\le b/\alpha\}\,dv\\
&=\frac{b-a}{\alpha}.
\end{align*}
Therefore the scalar disagreement probability is at most \(\Pr(N(U)\ge1)\le\mathbb E[N(U)]=(b-a)/\alpha\). Interchanging \(a\) and \(b\) gives the bound \(|a-b|/\alpha\) for arbitrary scalar values.

Applying this argument to each coordinate and taking a union bound yields
\[
\Pr\bigl(\mathsf Q_{\alpha,W}(v)\ne \mathsf Q_{\alpha,W}(v')\bigr)\le\sum_{\ell=1}^d\frac{|v_\ell-v'_\ell|}{\alpha}=\frac{\|v-v'\|_1}{\alpha}.
\]
Finally, \(\|v-v'\|_1\le\sqrt d\,\|v-v'\|_2\) by Cauchy--Schwarz. This proves both assertions.
\end{proof}

\begin{lemma}[Anytime GLM confidence]
\label{lem:repmaxinp-confidence}
With probability at least \(1-\eta\), \(\|\widehat\theta_t-\theta^\star\|_{V_t}\le\beta_t(\eta)\) simultaneously for every \(t\le T+1\). On the same event, \(\|\widetilde\theta_b-\theta^\star\|_{V_{\tau_b}}\le\Gamma_b\) at every realized batch.
\end{lemma}

\begin{proof}[Proof of Lemma~\ref{lem:repmaxinp-confidence}]
Let \(\mathcal G_\eta\) be the simultaneous event in Lemma~\ref{app:lem:repmaxinp-self-normalized}. It has probability at least \(1-\eta\). On this event, Lemma~\ref{app:lem:repmaxinp-estimation-bound} gives
\[
\|\widehat\theta_t-\theta^\star\|_{V_t}\le\frac{1}{c_\mu}\left[R\sqrt{\log\frac{\det(V_t)}{\det(\lambda I_d)}+2\log\frac1\eta}+\sqrt{\lambda}S\right]=\beta_t(\eta)
\]
simultaneously for every \(t\le T+1\).

Now consider any realized batch \(b\). The deterministic part of Lemma~\ref{lem:repmaxinp-grid} gives
\begin{align*}
\|\widetilde\theta_b-\widehat\theta_{\tau_b}\|_{V_{\tau_b}}
&=\left\|\mathsf Q_{\alpha_b,W_b}\bigl(V_{\tau_b}^{1/2}\widehat\theta_{\tau_b}\bigr)-V_{\tau_b}^{1/2}\widehat\theta_{\tau_b}\right\|_2\\
&\le\frac{\alpha_b\sqrt d}{2}\\
&=\beta_{\tau_b}(\eta)\frac{d}{\varrho-2\eta}.
\end{align*}
Consequently, on \(\mathcal G_\eta\),
\begin{align*}
\|\widetilde\theta_b-\theta^\star\|_{V_{\tau_b}}
&\le\|\widetilde\theta_b-\widehat\theta_{\tau_b}\|_{V_{\tau_b}}+\|\widehat\theta_{\tau_b}-\theta^\star\|_{V_{\tau_b}}\\
&\le\beta_{\tau_b}(\eta)\left(1+\frac{d}{\varrho-2\eta}\right)\\
&=\Gamma_b.
\end{align*}
The event \(\mathcal G_\eta\) controls every time, and the rounding bound holds for every shift. Hence these inequalities hold at every realized batch start, including when the batch starts depend on earlier observations.
\end{proof}

We will use the event
\begin{equation}\label{app:eq:repmaxinp-confidence-event}
\mathcal E:=\left\{\|\widetilde\theta_b-\theta^\star\|_{V_{\tau_b}}\le\Gamma_b\text{ for every realized batch }b\right\}.
\end{equation}
The preceding proof shows \(\mathcal G_\eta\subseteq\mathcal E\), so \(\Pr(\mathcal E)\ge1-\eta\ge1-\delta\). A single simultaneous confidence event covers all batches.

\subsection{Determinant growth and batch geometry}
\label{app:repmaxinp-geometry}

Let \(b(t)\) denote the batch containing round \(t\).

\begin{lemma}[Batch geometry]
\label{lem:repmaxinp-geometry}
On every trajectory, \(\log(\det(V_{T+1})/\det(\lambda I_d))\le D_{\max}\), at most \(B\) batches are started, and
\begin{equation}\label{eq:repmaxinp-geometry}
\|z_t\|_{V_{\tau_{b(t)}}^{-1}}\le\sqrt e\,\|z_t\|_{V_t^{-1}}, \qquad \sum_{t=1}^T\min\{1,\|z_t\|_{V_t^{-1}}^2\}\le2D_{\max}.
\end{equation}
\end{lemma}

\begin{proof}[Proof of Lemma~\ref{lem:repmaxinp-geometry}]
We prove the determinant bound, the bound on the number of batches, the comparison of uncertainty norms, and the summed uncertainty bound.

\paragraph{Global determinant bound.}
Since \(\|z_t\|_2\le2L\), we have \(\operatorname{tr}(V_{T+1})=\lambda d+\sum_{t=1}^T\|z_t\|_2^2\le\lambda d+4TL^2\). Applying the arithmetic--geometric mean inequality to the positive eigenvalues of \(V_{T+1}\) gives
\[
\det(V_{T+1})\le\left(\frac{\operatorname{tr}(V_{T+1})}{d}\right)^d\le\left(\lambda+\frac{4TL^2}{d}\right)^d.
\]
Dividing by \(\det(\lambda I_d)=\lambda^d\) and taking logarithms yields
\begin{equation}\label{app:eq:repmaxinp-determinant-bound}
\log\frac{\det(V_{T+1})}{\det(\lambda I_d)}\le d\log\left(1+\frac{4TL^2}{\lambda d}\right)=D_{\max}.
\end{equation}

\paragraph{Number of batches.}
Let \(J\) denote the number of batches started before the horizon ends. If \(J=1\), then \(J\le B\). Suppose \(J\ge2\). At each new batch start, the trigger gives \(\det(V_{\tau_{b+1}})>e\det(V_{\tau_b})\). Multiplying over the \(J-1\) triggers and using monotonicity of the determinant along the sequence \((V_t)\) gives
\[
\det(V_{T+1})\ge\det(V_{\tau_{J-1}})>e^{J-1}\det(V_1).
\]
Together with \eqref{app:eq:repmaxinp-determinant-bound}, this implies \(J-1<D_{\max}\). Since \(J-1\) is an integer, \(J\le\lceil D_{\max}\rceil\le B\). The strict trigger inequality gives this conclusion also when \(D_{\max}\) is an integer. The same argument applied to any prefix shows that the algorithm never requests a shift beyond \(W_{B-1}\).

\paragraph{Comparison of frozen and current norms.}
We first establish a matrix inequality. Let \(A\succeq C\succ0\) and define \(M:=C^{-1/2}AC^{-1/2}\). Every eigenvalue of \(M\) is at least one. Thus its largest eigenvalue is at most the product of its eigenvalues, giving \(M\preceq\det(M)I_d\). Inverting and conjugating by \(C^{-1/2}\) yields
\[
A^{-1}=C^{-1/2}M^{-1}C^{-1/2}\succeq\frac{1}{\det(M)}C^{-1}.
\]
Since \(\det(M)=\det(A)/\det(C)\), it follows that
\begin{equation}\label{app:eq:repmaxinp-matrix-comparison}
\|v\|_{C^{-1}}\le\sqrt{\frac{\det(A)}{\det(C)}}\,\|v\|_{A^{-1}}
\qquad\text{for every }v\in\mathbb R^d.
\end{equation}

Fix a round \(t\) in batch \(b=b(t)\). We have \(V_t\succeq V_{\tau_b}\). At the first round of the batch the determinant ratio is one. At any later round of the same batch, the preceding trigger test did not start a new batch, so \(\det(V_t)\le e\det(V_{\tau_b})\). Applying \eqref{app:eq:repmaxinp-matrix-comparison} with \(A=V_t\) and \(C=V_{\tau_b}\) gives
\begin{equation}\label{app:eq:repmaxinp-frozen-current}
\|z_t\|_{V_{\tau_b}^{-1}}\le\sqrt e\,\|z_t\|_{V_t^{-1}}.
\end{equation}
The comparison uses the matrix before the current observation, so it holds also at a round whose subsequent update triggers a new batch.

\paragraph{Summed uncertainty.}
Write \(a_t:=\|z_t\|_{V_t^{-1}}\). The matrix determinant lemma gives \(\det(V_{t+1})=\det(V_t)(1+a_t^2)\). For every \(u\ge0\), \(\log(1+u)\ge\frac12\min\{1,u\}\): for \(u\le1\) this follows by integrating \((1+s)^{-1}\ge1/2\) over \(s\in[0,u]\), and for \(u\ge1\) it follows from \(\log(1+u)\ge\log2\ge1/2\). Therefore,
\begin{align*}
\frac12\sum_{t=1}^T\min\{1,a_t^2\}
&\le\sum_{t=1}^T\log(1+a_t^2)\\
&=\log\frac{\det(V_{T+1})}{\det(V_1)}\\
&\le D_{\max}.
\end{align*}
This proves the second inequality in \eqref{eq:repmaxinp-geometry} and completes the lemma.
\end{proof}

We also record the consequence needed for regret. Since \(\min\{1,a\}^2=\min\{1,a^2\}\) for \(a\ge0\), Cauchy--Schwarz and Lemma~\ref{lem:repmaxinp-geometry} imply
\begin{equation}\label{app:eq:repmaxinp-linear-potential}
\sum_{t=1}^T\min\{1,\|z_t\|_{V_t^{-1}}\}\le\sqrt{T\sum_{t=1}^T\min\{1,\|z_t\|_{V_t^{-1}}^2\}}\le\sqrt{2TD_{\max}}.
\end{equation}

\subsection{Optimism and high-probability regret}
\label{app:repmaxinp-regret}

We first verify that pair selection is defined on every trajectory. Fix a round \(t\) in batch \(b\), and let \(\widehat i\) maximize \(x_{t,i}^\top\widetilde\theta_b\) over \(i\in[K_t]\). For every \(j\in[K_t]\), \((x_{t,\widehat i}-x_{t,j})^\top\widetilde\theta_b\ge0\). Since the confidence term is nonnegative, \(\widehat i\in\mathcal C_t\). Thus \(\mathcal C_t\) is nonempty, and the finite maximization over \(\mathcal C_t^2\) is well defined. This argument holds on every sample path.

\begin{lemma}[Optimism and instantaneous regret]
\label{lem:repmaxinp-optimism}
On \(\mathcal E\), every utility-maximizing index belongs to \(\mathcal C_t\), and
\begin{equation}\label{eq:repmaxinp-instantaneous}
r_t^{\mathrm{util}}\le2\Gamma_{b(t)}\|z_t\|_{V_{\tau_{b(t)}}^{-1}}.
\end{equation}
\end{lemma}

\begin{proof}[Proof of Lemma~\ref{lem:repmaxinp-optimism}]
Work on \(\mathcal E\) and fix a round \(t\) in batch \(b=b(t)\). For any vector \(v\in\mathbb R^d\), Cauchy--Schwarz gives
\begin{equation}\label{app:eq:repmaxinp-directional-confidence}
|v^\top(\widetilde\theta_b-\theta^\star)|\le\|v\|_{V_{\tau_b}^{-1}}\|\widetilde\theta_b-\theta^\star\|_{V_{\tau_b}}\le\Gamma_b\|v\|_{V_{\tau_b}^{-1}}.
\end{equation}

Let \(i^\star\) be any index maximizing \(x_{t,i}^\top\theta^\star\). For every \(j\in[K_t]\), optimality gives \((x_{t,i^\star}-x_{t,j})^\top\theta^\star\ge0\). Applying \eqref{app:eq:repmaxinp-directional-confidence} to \(v=x_{t,i^\star}-x_{t,j}\) gives
\begin{align*}
&(x_{t,i^\star}-x_{t,j})^\top\widetilde\theta_b+\Gamma_b\|x_{t,i^\star}-x_{t,j}\|_{V_{\tau_b}^{-1}}\\
&\qquad\ge(x_{t,i^\star}-x_{t,j})^\top\theta^\star\\
&\qquad\ge0.
\end{align*}
This holds for every \(j\), so \(i^\star\in\mathcal C_t\). Since \(i^\star\) was arbitrary among the maximizing indices, every optimal arm is promising.

Now fix the index \(i_t^\star\) used in the regret definition and write \(x_t^\star=x_{t,i_t^\star}\). For any \(i\in\mathcal C_t\), the defining inequality for \(\mathcal C_t\), evaluated against \(j=i_t^\star\), implies
\[
(x_t^\star-x_{t,i})^\top\widetilde\theta_b\le\Gamma_b\|x_t^\star-x_{t,i}\|_{V_{\tau_b}^{-1}}.
\]
Combining this with \eqref{app:eq:repmaxinp-directional-confidence} gives
\begin{align*}
(x_t^\star-x_{t,i})^\top\theta^\star
&=(x_t^\star-x_{t,i})^\top\widetilde\theta_b+(x_t^\star-x_{t,i})^\top(\theta^\star-\widetilde\theta_b)\\
&\le2\Gamma_b\|x_t^\star-x_{t,i}\|_{V_{\tau_b}^{-1}}.
\end{align*}

The indices \(i_t^\star,i_t,j_t\) all belong to \(\mathcal C_t\). The maximum-width selection rule therefore gives
\[
\|x_t^\star-x_t\|_{V_{\tau_b}^{-1}}\le\|x_t-y_t\|_{V_{\tau_b}^{-1}},
\qquad
\|x_t^\star-y_t\|_{V_{\tau_b}^{-1}}\le\|x_t-y_t\|_{V_{\tau_b}^{-1}}.
\]
Applying the preceding utility-gap bound to the two selected arms yields
\begin{align*}
r_t^{\mathrm{util}}
&=\frac12\left[(x_t^\star-x_t)^\top\theta^\star+(x_t^\star-y_t)^\top\theta^\star\right]\\
&\le\Gamma_b\left(\|x_t^\star-x_t\|_{V_{\tau_b}^{-1}}+\|x_t^\star-y_t\|_{V_{\tau_b}^{-1}}\right)\\
&\le2\Gamma_b\|x_t-y_t\|_{V_{\tau_b}^{-1}}\\
&=2\Gamma_b\|z_t\|_{V_{\tau_b}^{-1}}.
\end{align*}
This proves \eqref{eq:repmaxinp-instantaneous}.
\end{proof}

\begin{proof}[Regret bounds in Theorems~\ref{thm:repmaxinp} and~\ref{thm:repmaxinp-explicit}]
By \eqref{app:eq:repmaxinp-confidence-event}, the event \(\mathcal E\) has probability at least \(1-\delta\). We work on this event throughout the proof. Lemma~\ref{lem:repmaxinp-optimism} already gives the assertion that every optimal arm belongs to \(\mathcal C_t\) at every round.

First, we bound the confidence radii uniformly. Since \(V_{\tau_b}\preceq V_{T+1}\), Lemma~\ref{lem:repmaxinp-geometry} gives \(\log(\det(V_{\tau_b})/\det(\lambda I_d))\le D_{\max}\). Substituting \(\eta=\delta/B\) into \eqref{eq:repmaxinp-beta} yields
\[
\beta_{\tau_b}(\eta)\le\frac{1}{c_\mu}\left[R\sqrt{D_{\max}+2\log(B/\delta)}+\sqrt{\lambda}S\right].
\]
Also, \(d/(\varrho-2\eta)=dB/(\rho-2\delta)\). Hence \eqref{eq:repmaxinp-radii} and \eqref{eq:repmaxinp-gamma-bar} imply \(\Gamma_b\le\overline\Gamma\) at every realized batch.

Each selected arm has utility gap at most \(2LS\), since \(\|x_t^\star-x_{t,i}\|_2\le2L\) and \(\|\theta^\star\|_2\le S\). Thus \(r_t^{\mathrm{util}}\le2LS\). Combining this deterministic bound with Lemma~\ref{lem:repmaxinp-optimism}, and writing \(w_t:=\|z_t\|_{V_{\tau_{b(t)}}^{-1}}\), gives \(r_t^{\mathrm{util}}\le\min\{2LS,2\overline\Gamma w_t\}\).

Let \(M:=2\max\{LS,\overline\Gamma\}\). For every \(w\ge0\), \(\min\{2LS,2\overline\Gamma w\}\le M\min\{1,w\}\). Indeed, for \(w\le1\) the second term is at most \(Mw\), and for \(w\ge1\) the first term is at most \(M\). Using \eqref{app:eq:repmaxinp-frozen-current} and the inequality \(\min\{1,\sqrt e\,a\}\le\sqrt e\,\min\{1,a\}\) for \(a\ge0\), we obtain
\begin{align*}
R_T^{\mathrm{util}}
&\le M\sum_{t=1}^T\min\{1,w_t\}\\
&\le M\sqrt e\sum_{t=1}^T\min\{1,\|z_t\|_{V_t^{-1}}\}\\
&\le M\sqrt{2eTD_{\max}}\\
&=2\max\{LS,\overline\Gamma\}\sqrt{2eTD_{\max}}.
\end{align*}
The third inequality is \eqref{app:eq:repmaxinp-linear-potential}. This proves \eqref{eq:repmaxinp-hp}.

To obtain the simplified rate, treat \(L,S,R,\lambda\) as constants. Then \(D_{\max}=\widetilde O(d)\) and \(B=\widetilde O(d)\). Moreover, \(\rho-2\delta>\rho/3\), so the definition of \(\overline\Gamma\) gives
\[
\overline\Gamma=\widetilde O\left(\frac{1}{c_\mu}\left(\sqrt d+\frac{d^{5/2}}{\rho}\right)\right).
\]
Multiplication by \(\sqrt{TD_{\max}}=\widetilde O(\sqrt{dT})\) yields
\[
R_T^{\mathrm{util}}=\widetilde O\left(\frac{1}{c_\mu}\left(d+\frac{d^3}{\rho}\right)\sqrt T\right).
\]
The constant term \(LS\) in the maximum is absorbed by this bound because \(d\ge1\), \(c_\mu\le1\), and \(\rho<1\). Since \(d\ge1\) and \(\rho<1\), the term \(d\) is absorbed by \(d^3/\rho\). This proves \eqref{eq:repmaxinp-rate}.

For the expectation bound, let \(A:=2\max\{LS,\overline\Gamma\}\sqrt{2eTD_{\max}}\). The high-probability argument gives \(R_T^{\mathrm{util}}\le A\) on \(\mathcal E\), whereas \(R_T^{\mathrm{util}}\le2LST\) on every trajectory. Since \(\Pr(\mathcal E^c)\le\delta\),
\[
\mathbb E[R_T^{\mathrm{util}}]\le A\Pr(\mathcal E)+2LST\Pr(\mathcal E^c)\le A+2LST\delta.
\]
This proves \eqref{eq:repmaxinp-expectation}. The choice \(\delta=\min\{\rho/4,1/T\}\) satisfies \(0<\delta<\rho/3\), \(\rho-2\delta\ge\rho/2\), and \(T\delta\le1\). Thus the additional term is at most \(2LS\), and \(\log(1/\delta)=\log\max\{4/\rho,T\}\) contributes only a logarithmic factor. The expected rate in Theorem~\ref{thm:repmaxinp} follows.

\end{proof}

\subsection{Trajectory replicability}
\label{app:repmaxinp-coupling}

We now analyze the two executions under the shared-shift construction from Section~\ref{app:repmaxinp-setting}. The confidence bounds established above apply to each execution separately under its marginal law.

\paragraph{Histories before a batch shift.}
For each possible batch index \(b\in\{0,\ldots,B-1\}\), let \(\tau_b^{(k)}\) be the start of batch \(b\) in execution \(k\), if that batch is reached by time \(T\). To define the relevant joint history, stop each execution immediately before it would reveal \(W_b\), or at the horizon if it never reaches batch \(b\). Let \(\mathcal J_b\) be the sigma-field generated by these two stopped histories and the shifts \(W_0,\ldots,W_{b-1}\).

These stopped histories depend only on the earlier shifts and the comparison-noise streams. Neither stopped execution has used \(W_b\). Consequently, \(W_b\) remains independent of \(\mathcal J_b\) and has distribution \(\operatorname{Unif}([0,1)^d)\) conditional on \(\mathcal J_b\).

Let \(\mathcal M_b\) be the event that both executions reach batch \(b\) at the same time, with identical earlier ordered action pairs and agreement of all earlier rounded estimates. This is a \(\mathcal J_b\)-measurable event. When \(\mathcal M_b\) occurs, write \(\tau_b\) for the common start time and define
\begin{equation}\label{app:eq:repmaxinp-first-disagreement}
\mathcal L_b:=\mathcal M_b\cap\{\widetilde\theta_b^{(1)}\ne\widetilde\theta_b^{(2)}\}.
\end{equation}
The event \(\mathcal L_b\) is empty when such a common batch start is not reached.

\paragraph{Local confidence events.}
For each execution \(k\), define \(\mathcal E_b^{(k)}\) to be the event that its unrounded estimate satisfies the confidence bound at the start of batch \(b\):
\begin{equation}\label{app:eq:repmaxinp-local-confidence}
\left\|\widehat\theta_{\tau_b^{(k)}}^{(k)}-\theta^\star\right\|_{V_{\tau_b^{(k)}}^{(k)}}\le\beta_{\tau_b^{(k)}}^{(k)}(\eta).
\end{equation}
Take \(\mathcal E_b^{(k)}\) to hold when execution \(k\) does not reach batch \(b\). The radius on the right is computed from that execution's matrix. The event \(\mathcal E_b^{(k)}\) is \(\mathcal J_b\)-measurable because it involves only the observations and matrix available before the current rounding shift.

Let \(\mathcal G_\eta^{(k)}\) be the simultaneous unrounded confidence event in execution \(k\). Lemma~\ref{lem:repmaxinp-confidence} gives \(\Pr((\mathcal G_\eta^{(k)})^c)\le\eta\). Since \(\mathcal G_\eta^{(k)}\subseteq\mathcal E_b^{(k)}\), we have
\begin{equation}\label{app:eq:repmaxinp-local-failure}
\Pr\bigl((\mathcal E_b^{(k)})^c\bigr)\le\eta
\qquad\text{for every }b\text{ and }k\in\{1,2\}.
\end{equation}
Thus the local events are measurable before \(W_b\) is revealed, and their unconditional failure probabilities are controlled at the random invocation times.

\begin{lemma}[First batch disagreement]
\label{lem:repmaxinp-coupling}
For every \(b\in\{0,\ldots,B-1\}\),
\begin{equation}\label{eq:repmaxinp-batch-disagreement}
\Pr(\mathcal L_b)\le\varrho=\rho/B.
\end{equation}
\end{lemma}

\begin{proof}[Proof of Lemma~\ref{lem:repmaxinp-coupling}]
Fix a batch index \(b\). On \(\mathcal M_b\), the two executions have selected the same earlier ordered action pairs from the same indexed feature sequence. Their comparison features therefore agree through the common time \(\tau_b-1\), and their design matrices coincide. Write this common matrix as \(V\). The radius \(\beta_{\tau_b}(\eta)\) depends on these histories only through \(V\), so it also agrees; write it as \(\beta\). The grid widths agree and equal \(\alpha=2\beta\sqrt d/(\varrho-2\eta)\).

On \(\mathcal M_b\), define the two vectors \(v^{(k)}:=V^{1/2}\widehat\theta_{\tau_b}^{(k)}\). Conditional on \(\mathcal J_b\), these vectors and the grid width are fixed. On the \(\mathcal J_b\)-measurable event \(\mathcal M_b\cap\mathcal E_b^{(1)}\cap\mathcal E_b^{(2)}\), the triangle inequality gives
\begin{align*}
\|v^{(1)}-v^{(2)}\|_2
&=\|\widehat\theta_{\tau_b}^{(1)}-\widehat\theta_{\tau_b}^{(2)}\|_V\\
&\le\|\widehat\theta_{\tau_b}^{(1)}-\theta^\star\|_V+\|\widehat\theta_{\tau_b}^{(2)}-\theta^\star\|_V\\
&\le2\beta.
\end{align*}
The two executions round using the same fresh \(W_b\). Since multiplication by the common \(V^{-1/2}\) is invertible, disagreement of the rounded parameter vectors is equivalent to disagreement of their quantized vectors. Lemma~\ref{lem:repmaxinp-grid} therefore implies, on this event,
\begin{align*}
\Pr\bigl(\widetilde\theta_b^{(1)}\ne\widetilde\theta_b^{(2)}\mid\mathcal J_b\bigr)
&\le\frac{\sqrt d\,\|v^{(1)}-v^{(2)}\|_2}{\alpha}\\
&\le\frac{2\sqrt d\,\beta}{\alpha}\\
&=\varrho-2\eta.
\end{align*}

Decompose \(\mathcal L_b\) according to whether the two local confidence events hold. Using their measurability, the tower property, and \eqref{app:eq:repmaxinp-local-failure}, we obtain
\begin{align*}
\Pr(\mathcal L_b)
&\le\Pr\bigl((\mathcal E_b^{(1)})^c\bigr)+\Pr\bigl((\mathcal E_b^{(2)})^c\bigr)\\
&\quad+\mathbb E\left[\mathbf 1_{\mathcal M_b\cap\mathcal E_b^{(1)}\cap\mathcal E_b^{(2)}}\Pr\bigl(\widetilde\theta_b^{(1)}\ne\widetilde\theta_b^{(2)}\mid\mathcal J_b\bigr)\right]\\
&\le2\eta+(\varrho-2\eta)\Pr\bigl(\mathcal M_b\cap\mathcal E_b^{(1)}\cap\mathcal E_b^{(2)}\bigr)\\
&\le2\eta+(\varrho-2\eta)\\
&=\varrho.
\end{align*}
This proves \eqref{eq:repmaxinp-batch-disagreement}. The conditional step uses only the fresh shared shift; the statistical failure probabilities enter through the unconditional bounds in \eqref{app:eq:repmaxinp-local-failure}.
\end{proof}

\begin{proof}[Proof of the replicability assertion in Theorem~\ref{thm:repmaxinp}]
The bound of \(B\) on the number of batches was established in Lemma~\ref{lem:repmaxinp-geometry}. It remains to prove agreement of the ordered action trajectories.

Both executions begin batch zero at time one with matrix \(\lambda I_d\). Thus \(\mathcal M_0\) holds. We show that whenever \(\mathcal M_b\) holds and \(\mathcal L_b\) does not, the executions select the same pairs throughout batch \(b\) and either both terminate at the horizon or reach the next batch in the event \(\mathcal M_{b+1}\).

On \(\mathcal M_b\cap\mathcal L_b^c\), the two rounded estimates agree. The matrices at the common batch start agree because their earlier comparison-feature sequences agree. Their confidence radii, being deterministic functions of these matrices and the common budgets, also agree. Hence both executions have identical frozen quantities \(\widetilde\theta_b\), \(\Gamma_b\), and \(V_{\tau_b}\).

At the common first round of this batch, both executions observe the same indexed feature set. The definition of \(\mathcal C_t\) therefore produces the same set of promising indices. The objective in the maximum-width selection rule is the same for every candidate ordered pair, and lexicographic tie-breaking produces the same \((i_t,j_t)\). This conclusion holds also when some features or objective values coincide.

Because the selected indices and contexts agree, the comparison feature \(z_t\) agrees. Both executions consequently update their design matrices by the same rank-one matrix \(z_tz_t^\top\). Their observed outcomes may differ, but these outcomes are retained only for the next estimator call. They do not alter the frozen quantities used to select pairs within the current batch.

The determinant test is a deterministic function of the current matrix and the frozen batch-start matrix. It therefore has the same result in both executions. If it does not trigger and the horizon has not ended, the preceding argument applies at the next round. If it triggers, both executions start batch \(b+1\) at the same time. Their earlier action pairs agree and all rounded estimates through batch \(b\) agree, so \(\mathcal M_{b+1}\) holds. If the horizon ends, their action pairs agree through time \(T\). This proves the claimed implication by induction within the batch.

Applying this implication successively from batch zero shows that, outside \(\bigcup_{b=0}^{B-1}\mathcal L_b\), the complete ordered action trajectories agree. Equivalently,
\begin{equation}\label{app:eq:repmaxinp-trajectory-inclusion}
\left\{\exists t\in[T]:(i_t^{(1)},j_t^{(1)})\ne(i_t^{(2)},j_t^{(2)})\right\}\subseteq\bigcup_{b=0}^{B-1}\mathcal L_b.
\end{equation}
Taking probabilities and applying Lemma~\ref{lem:repmaxinp-coupling} gives
\begin{align*}
\Pr\left(\exists t\in[T]:(i_t^{(1)},j_t^{(1)})\ne(i_t^{(2)},j_t^{(2)})\right)
&\le\sum_{b=0}^{B-1}\Pr(\mathcal L_b)\\
&\le B\varrho\\
&=\rho.
\end{align*}
This is trajectory replicability in the sense of Definition~\ref{def:replicability}. Together with the regret proof, it completes Theorems~\ref{thm:repmaxinp} and~\ref{thm:repmaxinp-explicit}.
\end{proof}

\subsection{Initialization and the relation between regret measures}
\label{app:repmaxinp-remarks}

We record initialization details and relate utility regret to the probability regret used elsewhere in the paper.

\paragraph{Initialization.}
At \(t=1\), the sum in \eqref{eq:repmaxinp-estimator} is empty, so the objective is \(-\lambda\|\theta\|_2^2/2\). Since \(0\in\Theta\) and \(\lambda>0\), its unique maximizer is \(\widehat\theta_1=0\). Also, \(V_1=\lambda I_d\succ0\), so the square roots, inverse matrices, and uncertainty norms are defined from the first round.

\paragraph{Binary-noise parameter.}
Conditional on \(\mathcal F_t^{-}\), the variable \(O_t\) takes values in \([0,1]\) and has mean \(\mu(z_t^\top\theta^\star)\). Hoeffding's lemma therefore gives \(\mathbb E[\exp(a\varepsilon_t)\mid\mathcal F_t^{-}]\le\exp(a^2/8)\) for every \(a\in\mathbb R\). This is the sub-Gaussian condition with \(R=1/2\).

\paragraph{Logistic curvature.}
For \(\mu(u)=(1+e^{-u})^{-1}\), the derivative is
\[
\mu'(u)=\mu(u)(1-\mu(u))=\frac{1}{2+e^u+e^{-u}}.
\]
This is an even function that decreases as \(|u|\) increases. Since \(|z^\top\theta|\le2LS\) whenever \(\|z\|_2\le2L\) and \(\theta\in\Theta\), it follows that \(\mu'(z^\top\theta)\ge\mu(2LS)(1-\mu(2LS))\). This quantity lies in \((0,1]\) and is a valid choice of \(c_\mu\).

\paragraph{Probability regret.}
The symmetry of the link gives \(\mu(0)=1/2\). For any available arm \(x\) at round \(t\), define its utility gap \(a:=(x_t^\star-x)^\top\theta^\star\). Optimality and boundedness give \(0\le a\le2LS\). Let \(z:=x_t^\star-x\), so \(\|z\|_2\le2L\). For every \(q\in[0,1]\), the vector \(qz\) also has norm at most \(2L\), and therefore \(\mu'(qa)=\mu'((qz)^\top\theta^\star)\ge c_\mu\). By continuity, \(k_\mu:=\sup_{|u|\le2LS}\mu'(u)\) is finite. Integrating the derivative along the segment from zero to \(a\) gives
\begin{equation}\label{app:eq:repmaxinp-link-gap}
\mu(a)-\frac12=a\int_0^1\mu'(qa)\,dq,
\qquad
c_\mu a\le\mu(a)-\frac12\le k_\mu a.
\end{equation}
These inequalities also hold when \(a=0\).

Apply \eqref{app:eq:repmaxinp-link-gap} to the utility gaps \(a_t^x:=(x_t^\star-x_t)^\top\theta^\star\) and \(a_t^y:=(x_t^\star-y_t)^\top\theta^\star\). The instantaneous probability regret is
\[
r_t=\frac12\left[\mu(a_t^x)-\frac12+\mu(a_t^y)-\frac12\right],
\]
while \(r_t^{\mathrm{util}}=(a_t^x+a_t^y)/2\). Hence \(c_\mu r_t^{\mathrm{util}}\le r_t\le k_\mu r_t^{\mathrm{util}}\). Summing over rounds yields
\begin{equation}\label{app:eq:repmaxinp-regret-comparison}
c_\mu R_T^{\mathrm{util}}\le R_T\le k_\mu R_T^{\mathrm{util}}.
\end{equation}
In particular, on the event in Theorem~\ref{thm:repmaxinp}, the probability regret is bounded by \(k_\mu\) times the right-hand side of \eqref{eq:repmaxinp-hp}.

%% file: appendix/experimental_details.tex
\subsection{Synthetic Preferences}
\label{sec:app-synthetic}

\paragraph{Instances and protocol.} We use Bradley--Terry--Luce instances with $K=5$ arms and $P_{i,j}=\sigma(\theta_i-\theta_j)$. The best arm has utility $0$, and the remaining utilities are $-\Delta_i$, with $\Delta_i\sim U[0.3,2.5]$ independently; arm labels are randomly permuted. These instances satisfy Assumptions~3.1--3.4 with $\gamma=1$. Paired executions share the algorithm's opponent draws, tie-breaking randomness, and rounding offsets, as applicable, but receive independent comparison outcomes. Counter-based generators keyed by seed, purpose, and index preserve the shared randomness across executions. We report $F(t)$, the fraction of pairs with identical ordered comparison sequences through round $t$. Evaluation uses $30$ test instances disjoint from the $12$ validation instances, with $200$ paired runs per test instance and fresh seeds. Means and standard errors are computed across test instances.

\paragraph{Algorithms and tuning.} Confidence widths are tuned as $\kappa/\sqrt{n}$. BTM uses $c^*=\kappa/\sqrt{n^*}$. RepBTM uses $\tau(q)=\kappa/\sqrt q$, rounding width $\tau(q)/2$, and elimination threshold $6\gamma^2\tau(q)$; RepBTM-SST uses $\tau_g=\kappa/\sqrt{n_g}$, rounding width $\tau_g/2$, and threshold $4\tau_g$. We search geometric grids with ratio $2^{1/4}$ over $[2^{-2},2^4]$, $[1,2^{6.25}]$, and $[1,2^{6.75}]$ for BTM, RepBTM, and RepBTM-SST, respectively. Each value receives $100$ paired runs per validation instance; for each $\rho$, the three lowest-regret admissible values and the grid neighbours of the best receive $200$ additional pairs per instance. Admissibility requires $F_{\mathrm{lo}}\ge1-\rho$, where $F_{\mathrm{lo}}$ is the smaller of the one-sided $95\%$ Clopper--Pearson lower bound on pooled agreement and an instance-clustered $t$ lower bound. We select the lowest-regret value that is admissible together with every larger grid value and eliminates at least one arm on average. BTM is tuned on regret alone.

\paragraph{Results.} The horizon is $T=2.5\times10^8$. The regret and agreement curves use $\rho=0.2$ and display the first $6\times10^6$ rounds, by which all selected test runs have committed to the best arm; the curves remain flat thereafter. The right panel reports final regret for $\rho\in\{0.1,0.2,0.3,0.4,0.5\}$. Bands and error bars show $95\%$ intervals across the $30$ test instances. Reported agreement is averaged across instances.

\subsection{Preferences from a Language Model}
\label{app:llm}

\paragraph{Preference data.} Qwen3.8-27B generates summaries of $30$ Wikipedia articles (CC BY-SA~4.0) for three audiences, namely, a child, a humanities student, and a domain expert, at two target lengths, either $30$ or $100$-to-$250$ words. These are our $K=6$ arms. A different instance of the same model judges pairs while simulating a humanities student who prefers longer summaries. For each article, each pair of summaries is judged $30$ times, equally divided between the two display orders. Empirically, the arm $w=(\text{student},\text{long})$ is the Condorcet winner for this setting, which also matches the ground truth in our design. Prompts, preference matrices, and code are provided in the supplementary material.

\paragraph{Protocol.} Round $t$ uses article $c_t=t\bmod30$, and comparing summaries $(i,j)$ returns a uniformly sampled response from the already computed bank of $30$ comparisons between $i$ and $j$ for the article $c_t$. The instantaneous regret is $r_t=\frac12(\epsilon^{c_t}_{i_t}+\epsilon^{c_t}_{j_t})$, where $\epsilon^c_i=P^c_{w,i}-\frac12$, where $P^c_{w,i}$ measures the probability with which the Condorcet arm $w$ beats arm $i$ for the article $c$. Paired executions share the article sequence, opponent draws, and rounding offsets, and use independent comparison outcomes. We report $F(t)$, the fraction of paired executions whose ordered comparison sequences agree through round $t$.

\paragraph{Algorithms.} The experimental RepBTM-SST implementation uses the phase schedule of Algorithm~2 with rounding width $h_g=c/\sqrt{n_g}$. Each phase mean is mapped to the midpoint of its shared grid cell intersected with $[0,1]$, and arm $i$ is eliminated when $\max_j\widehat p_{j,g}-\widehat p_{i,g}>2a\big(\sqrt{\log(2K(g+1)^2)/(2n_g)}+h_g/2\big)$. Setting $c=0$ gives the unrounded batched baseline. The sequential baseline cycles leaders over the active set, samples opponents uniformly, retains observations across eliminations, and updates after every comparison. It removes every arm whose upper confidence bound lies below the largest lower confidence bound, using radius $a\sqrt{\log(2Kn_i^2)/(2n_i)}$ after $n_i$ observations led by arm $i$. Both baselines use $a=1$.

\textbf{Model and Compute.} Summaries and judgments come from Qwen3.8-27B with 8-bit weights and thinking
disabled, served by llama.cpp on one RTX 5090 GPU. Summaries are sampled at
temperature 0.2 and judgments at 0.7, both with top-p 0.8 and top-k 20.

\paragraph{Tuning and Results.} A configuration is eligible if it eliminates $w$ in at most $5\%$ of executions and ends with at most $1.05$ active arms on average. Among eligible configurations with empirical agreement $F(T)\ge1-\rho$, we select the one with the lowest mean regret. The left and middle panels use $T=3\times10^4$, $\rho=0.25$, and $200$ paired runs; the selected parameters are $(c,a)=(16,0.5)$. The right panel uses a larger horizon of $T=3\times10^5$. Figure~\ref{fig:llm} provides evaluation agreement rates below their targets; shaded bands indicate $95\%$ intervals across paired runs.
